\documentclass[acmtog]{acmart}
\AtBeginDocument{%
  }

\acmSubmissionID{611s1}

\newcommand{\tabfirst}[1]{\textbf{#1}}
\newcommand{\tabsecond}[1]{\underline{#1}}

\usepackage{multirow}
\usepackage{makecell}
\usepackage{enumitem}
\usepackage[percent]{overpic}
\usepackage{lipsum}
\usepackage{annotate-equations}

\let\titleold\title
\renewcommand{\title}[1]{\titleold{#1}\newcommand{\thetitle}{#1}}
\def\maketitlesupplementary
   {
   \newpage
       \twocolumn[
        \centering
        \Large
        \textbf{\thetitle}\\
        \vspace{0.5em}Appendix \\
        \vspace{1.0em}
       ] 
   }

\begin{document}

\title{MyTimeMachine: Personalized Facial Age Transformation}

\author{Luchao Qi}
\affiliation{%
  \institution{University of North Carolina at Chapel Hill}
  \city{Chapel Hill}
  \state{North Carolina}
  \country{USA}}
\email{lqi@cs.unc.edu}

\author{Jiaye Wu}
\affiliation{%
  \institution{University of Maryland}
  \city{College Park}
  \state{Maryland}
  \country{USA}}
\email{jiayewu@cs.umd.edu}

\author{Bang Gong}
\email{gongbang@cs.unc.edu}
\author{Annie N. Wang}
\email{awang13@cs.unc.edu}
\affiliation{%
  \institution{University of North Carolina at Chapel Hill}
  \city{Chapel Hill}
  \state{North Carolina}
  \country{USA}}

\author{David W. Jacobs}
\affiliation{%
  \institution{University of Maryland}
  \city{College Park}
  \state{Maryland}
  \country{USA}}
\email{dwj@umd.edu}

\author{Roni Sengupta}
\affiliation{%
  \institution{University of North Carolina at Chapel Hill}
  \city{Chapel Hill}
  \state{North Carolina}
  \country{USA}}
\email{ronisen@cs.unc.edu}

\renewcommand{\shortauthors}{Qi et al.}

\begin{abstract}
Facial aging is a complex process, highly dependent on multiple factors like gender, ethnicity, lifestyle, etc., making it extremely challenging to learn a global aging prior to predict aging for any individual accurately. Existing techniques often produce realistic and plausible aging results, but the re-aged images often do not resemble the person's appearance at the target age and thus need personalization. In many practical applications of virtual aging, e.g. VFX in movies and TV shows, access to a personal photo collection of the user depicting aging in a small time interval (20$\sim$40 years) is often available. However, naive attempts to personalize global aging techniques on personal photo collections often fail. 
Thus, we propose MyTimeMachine (MyTM), a method that combines a global aging prior with a personalized photo collection (ranging from as few as 10 images, ideally 50) to learn individualized age transformations.
We introduce a novel Adapter Network that combines personalized aging features with global aging features and generates a re-aged image with StyleGAN2. We also introduce three loss functions to personalize the Adapter Network with personalized aging loss, extrapolation regularization, and adaptive w-norm regularization. Our approach can also be extended to videos, achieving high-quality, identity-preserving, and temporally consistent aging effects that resemble actual appearances at target ages, demonstrating its superiority over state-of-the-art approaches.
\end{abstract}




\begin{teaserfigure}
\vspace{2.5em}
\begin{overpic}[width=\linewidth]{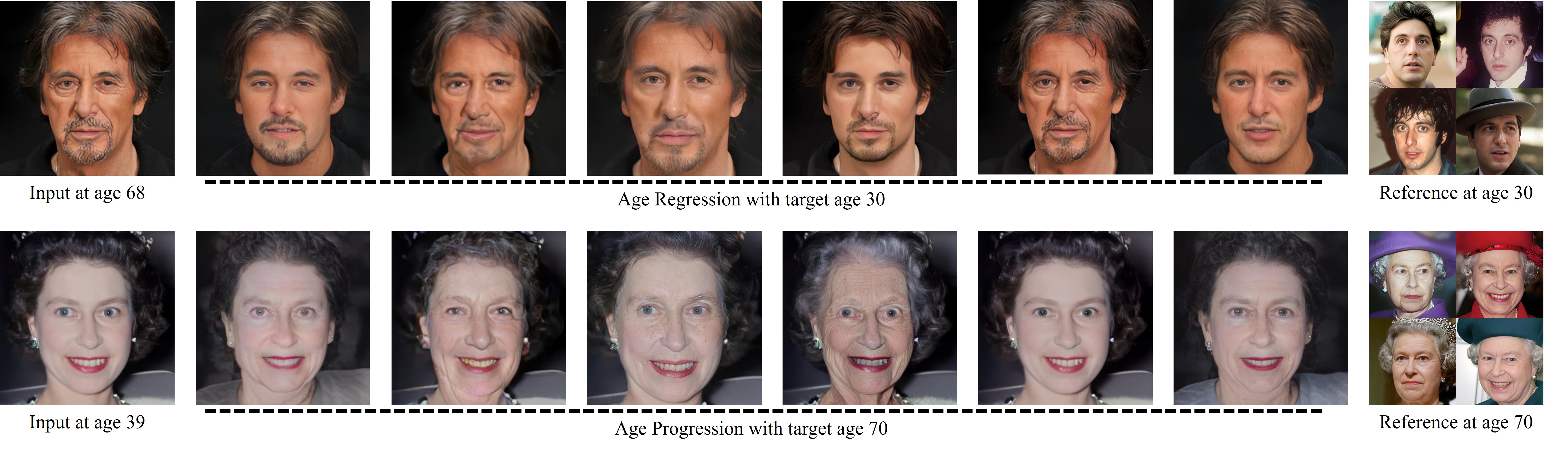}
\put(17,   30.4){\scriptsize SAM}
\put(13.6,   29.2){\scriptsize\cite{alaluf_only_2021}}
\put(29,   30.4){\scriptsize CUSP}
\put(23.5,  29.2){\scriptsize \cite{gomez-trenado_custom_2022}}
\put(39.5, 30.4){\scriptsize AgeTransGAN}
\put(39.1, 29.2){\scriptsize \cite{avidan_agetransgan_2022}}
\put(53.5, 30.4){\scriptsize FADING}
\put(48.7, 29.2){\scriptsize \cite{Chen_2023_BMVC}}
\put(64, 30.4){\scriptsize + DreamBooth}
\put(63.9, 29.2){\scriptsize \cite{ruiz_dreambooth_2023}}
\put(76.3, 29.2){\scriptsize MyTimeMachine}
\end{overpic}
\captionof{figure}{
We introduce MyTimeMachine (MyTM) to perform personalized age regression (top) and progression (bottom) by training a person-specific aging model from a few (10$\sim$50) personal photos spanning over a 20-40 year range. Our method outperforms existing age transformation techniques to generate re-aged faces that closely resemble the characteristic facial appearance of the user at the target age.
\label{fig:teaser}
}
\end{teaserfigure}

\setcopyright{acmlicensed}
\acmJournal{TOG}
\acmYear{2025} \acmVolume{44} \acmNumber{4} \acmArticle{} \acmMonth{8}\acmDOI{10.1145/3731172}

\maketitle

\section{Introduction}
\label{sec:intro}
What makes face aging so challenging? Virtual age transformation algorithms aim to digitally simulate the physical aging process of an individual's face. The goal of these methods \cite{zhang_age_2017, he_s2gan_2019, or-el_lifespan_2020, makhmudkhujaev_re-aging_2021, li_continuous_2021, avidan_agetransgan_2022, xie_diverse_2024} is to modify the shape and texture of the face to create the desired re-aging effect, while preserving the individual's unique identity, along with the pose, lighting, and style of the input image. 
However, facial aging is often highly dependent on several factors, such as ethnicity, gender, genetics, lifestyle, and health conditions \cite{swift_facial_2020, mendelson_changes_2012}, which makes it challenging to model.

Existing age transformation algorithms \cite{alaluf_only_2021, avidan_agetransgan_2022, gomez-trenado_custom_2022} learn a generative global prior, modeling how an average face ages, typically using datasets like FFHQ~\cite{karras_style-based_2019}. While these methods generate visually appealing results, they often fail to reflect how a specific individual actually ages. For example, given an image of \textit{Al Pacino} at 68, state-of-the-art techniques \cite{alaluf_only_2021, gomez-trenado_custom_2022, avidan_agetransgan_2022, Chen_2023_BMVC} produce a realistic but inaccurate version of his 30-year-old self (see Fig.~\ref{fig:teaser}). In applications like re-aging actors, accurate personalization is essential, as audiences are often familiar with a subject’s appearance over time.
However, accurately predicting an individual's re-aged appearance from a single image is highly ill-posed and challenging, since aging is person-specific~\cite{despois_agingmapgan_2020}.

In this paper, we show that accurate age synthesis can be achieved with as few as 10, ideally 50, photos of an individual across a 20$\sim$40 year time range. Personal photo collections are often available in many practical applications of virtual aging, and utilizing them can significantly improve the result, see MyTimeMachine in Fig~\ref{fig:teaser}. For example, consider de-aging effects often used in movies where a particular actor at 60 years old is shooting a scene where they need to be rendered as 30 years old. We can easily access the past 20$\sim$40 years of photos of the actor to learn an accurate aging model. Similarly, consider an individual interested in simulating how a photograph of their loved one at 40 years would appear at 60 years old or beyond. We can also easily access the past 10-20 years of photo collection of their loved ones to understand the aging process and more accurately simulate their future appearance at 60 years and beyond. 
We therefore create a personalized aging method that can transform an input image to any target age, both within and beyond the age range represented in the personal photo collection used for training.

Simply personalizing a generic global age transformation algorithm, e.g. FADING~\cite{Chen_2023_BMVC}, with Dreambooth~\cite{ruiz_dreambooth_2023} is ineffective.
Personal photo collections often cover a limited range of age, pose, lighting, and style variations compared to large-scale facial datasets like FFHQ. Consequently, naive fine-tuning typically results in overfitting, limiting the model's ability to generalize to unseen ages, poses, styles, and lighting conditions, as shown by the extrapolation failure of FADING + Dreambooth in row 4 of Fig.~\ref{fig:age_reg}. 
Additionally, FADING is built on diffusion, facing the typical inversion-editability trade-off problem~\cite{tov_designing_2021,hertz_prompt--prompt_2022}. Specifically, re-aging requires both high fidelity to the input face at similar ages and high editability as the target age diverges. In contrast, such trade-offs have been more well explored in StyleGAN2's well-trained latent space~\cite{roich_pivotal_2021, tov_designing_2021, bhattad_make_2023, xia_gan_2022}. Therefore, we demonstrate an effective approach to personalized age transformation based on StyleGAN2.

Our proposed personalized age transformation network, MyTimeMachine (MyTM), introduces a novel adapter network that updates global facial aging features with personalized aging characteristics, trained on a personal photo collection using custom loss functions. Built on top of SAM \cite{alaluf_only_2021}, a global age transformation network capable of continuous aging without per-image optimization, MyTM enhances SAM’s global age encoder, which projects an input image into StyleGAN2's latent space with a specified target age. We design a personalized adapter network that learns to adjust the global aging features. To train this adapter, we introduce three loss functions: personalized aging loss, extrapolation regularization, and adaptive w-norm regularization. The personalized aging loss ensures that identity-preserving features of the reaged image closely resemble those in a reference image from the personal photo collection at a similar target age. Extrapolation regularization controls aging effects beyond the training age range using global priors, while adaptive w-norm regularization addresses StyleGAN's inversion-editability trade-off, ensuring distinct shape and texture changes due to aging while preserving identity. We extend MyTM to video re-aging by utilizing face-swapping techniques to generate temporally consistent and identity-preserving results.



We curated a longitudinal aging dataset comprising high-quality images of 12 celebrities, captured under diverse conditions, including varying poses, expressions, and lighting. Inspired by real-world applications of personalized aging, we train our model on this dataset and establish two experimental frameworks to evaluate its performance: one for age regression, where a 70-year-old is rendered younger, and another for age progression, where a 40-year-old is rendered older. Our method outperforms existing global age transformation and naive personalization techniques, delivering high-quality, identity-preserving aging effects in both images and videos that resemble individual’s appearance at the target age.


In summary, our contributions are as follows: (i) We demonstrate that with access to a few ($\sim$50) personal images spanning a few decades (20$\sim$40 years), we can achieve high-quality, identity-preserving facial age transformations. These transformations accurately reflect the person's appearance at the target age while maintaining the style of the input image. (ii) We introduce several key technical advancements that integrate a global aging prior with a personal photo collection to enable personalized aging. Our approach trains an adapter network to adjust the global aging prior, utilizing three custom loss functions: personalized age loss, extrapolation regularization, and adaptive w-norm regularization. (iii) We show that MyTM can also be extended to perform temporally consistent and identity-preserving reaging in videos, which is important for many VFX applications.


\section{Related Work}
\label{sec:related_work}

Traditional age transformation methods fall into two categories: prototype-based \cite{tiddeman_prototyping_2001, kemelmacher-shlizerman_illumination-aware_2014} and physical model-based approaches \cite{suo_compositional_2010, tazoe_facial_2012}. For a detailed overview, we refer readers to the survey by \cite{aging_survey}. Recently, generative models have shown impressive results in synthesizing and editing high-resolution face images, inspiring their use in aging tasks.

\subsection{Global Age Editing}
Global age editing refers to age transformation without personal data. Prior works \cite{shen_interfacegan_2020, nitzan_large_2022} leverage StyleGAN2's well-trained latent space, identifying and traversing a linear age editing direction within it.
However, this assumption often fails with larger age changes, especially across a lifespan, and can entangle other attributes (e.g., gender or glasses) \cite{harkonen_ganspace_2020}. To address this, recent methods~\cite{makhmudkhujaev_re-aging_2021, or-el_lifespan_2020, yao_high_2020} apply nonlinear latent age transformations by training dedicated age encoders.

In diffusion models, recent method~\cite{wahid2024diffage3d} combines diffusion with EG3D~\cite{chan_efficient_2022} for 3D age editing.  Other methods~\cite{li_pluralistic_2023, baumann_continuous_2024, kwon_diffusion_2023, 10810706} perform 2D age editing through latent manipulation guided by CLIP~\cite{radford_learning_2021}. However, these methods continue to struggle with attribute entanglement in the latent space. FADING~\cite{Chen_2023_BMVC} improves disentanglement by projecting the input face into the diffusion model’s latent space using NTI~\cite{mokady_null-text_2022} and applying age editing through p2p~\cite{hertz_prompt--prompt_2022}.
However, FADING focuses on textural changes to achieve re-aging, neglecting the broader facial shape changes that occur over a person’s lifespan~\cite{gomez-trenado_custom_2022}. This limitation arises because p2p identifies age-related pixels through attention control, resulting in localized edits rather than facial structural changes~\cite{rout_semantic_2024}.
To address this, we build on StyleGAN2, leveraging its well-trained latent space for both fine-grained textural control and structural changes.


\subsection{Personalization of Generative Models}
Personalization involves tuning face models with personal images. PTI~\cite{roich_pivotal_2021} fine-tunes a StyleGAN2 generator anchored by an inverted latent code. Other approaches~\cite{nitzan_mystyle_2022, zeng_mystyle_2023, qi_my3dgen_2023} adapt the generator on a small set of personal images (50$\sim$100) to create a personalized prior.
In diffusion models, Dreambooth~\cite{ruiz_dreambooth_2023} optimizes the weights of the network to adapt to a specific subject through a prompt identifier~\cite{10504891}. 

In the context of lifespan age transformation, these personalization techniques often overfit to the few training images in a limited range (e.g., ages 50 to 70), making it challenging to extrapolate to ages beyond the training range (e.g., 20 years old). We demonstrate that MyTM produces personalized face aging results within the training age range and generalizes to ages beyond it.

\subsection{Video Re-aging}
A recent video re-aging approach, FRAN~\cite{zoss_production-ready_2022}, applies facial masks to predict age-related changes within masked regions per frame.
However, similar to FADING, such method often overlooks structural changes in facial shape that naturally occur over a person’s lifetime, such as the widening of a previously narrow face due to bone growth and shifts in facial fat distribution with age~\cite{gomez-trenado_custom_2022}. It also suffers with temporal consistency since it is trained on static images. To address this, Muqeet et al.~\cite{muqeet_video_2023} propose a re-aging model trained on synthetic video data, generating re-aged keyframes and interpolating between them to enhance temporal consistency. 
However, neither of these approaches is open-sourced or supports personalized video re-aging.
Recent work~\cite{iperov_iperovdeepfacelive_2024} seeks to enhance temporal consistency in identity-specific face-swapping by personalizing models for individual users. However, this approach demands around 5,000 images of the person’s face captured under various conditions, all at a similar age, limiting its effectiveness for lifespan aging transformations.

To address this, we follow face-swapping techniques~\cite{chen_simswap_2021,xu_mobilefaceswap_2022}, using our personalized re-aged face as the source for swapping. This approach eliminates the need for training a dedicated model on a large number of personal images.

\begin{figure*}[t]
    \centering
    \begin{overpic}[width=\linewidth]{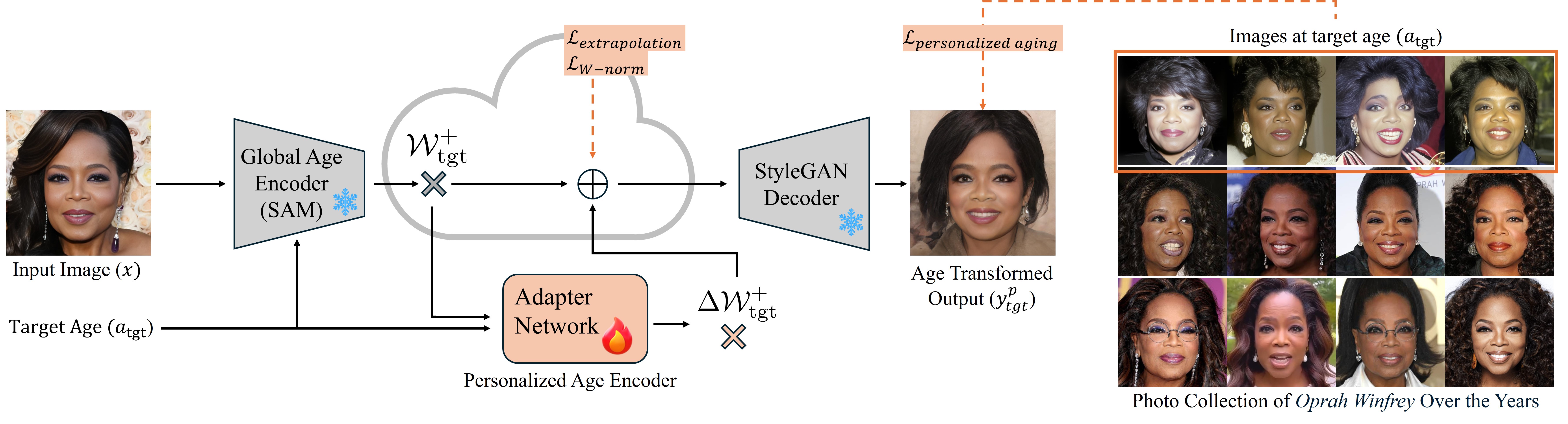}

    \put(44, 24.3){\scriptsize Eq.~\ref{eq:reg_extrapolation}}
    \put(41.5, 22.5){\scriptsize Eq.~\ref{eq:reg_w_norm}}
    \put(68, 24.3){\scriptsize Eq.~\ref{eq:pers_aging_loss}}
    \end{overpic}
    \caption{
    Given an input face of \textit{Oprah Winfrey} at 70 years old, our adapter re-ages her face to resemble her appearance at 30, while preserving the style of the input image.
    To achieve personalized re-aging, we collect $\sim$50 images of an individual across different ages and train an adapter network that updates the latent code generated by the global age encoder SAM.
    Our adapter preserves identity during interpolation when the target age falls within the range of ages seen in the training data, while also extrapolating well to unseen ages.} 
    \label{fig:architecture}
\end{figure*}

\section{Method}
In Section~\ref{sec:prelim}, we outline the preliminaries of global aging—SAM \cite{alaluf_only_2021}. Section~\ref{sec:design} introduces our personalized aging adapter, MyTM, and how it combines personal and global aging. Finally, in Section~\ref{sec:loss}, we describe the training losses for MyTM, explaining personalized re-aging for both interpolation and extrapolation (target ages within or beyond the training range).

\subsection{Preliminaries}
\label{sec:prelim}
Here we provide a brief overview of SAM~\cite{alaluf_only_2021}, a global age transformation network that forms the building block of our proposed personalized network, MyTM. SAM trains an age encoder ($\operatorname{E}_{\theta}$) that maps an input image ($x$) into the latent space $\mathcal{W}^+$ of StyleGAN, aligning with the desired target age ($a_\text{tgt}$). The latent code is then processed through the pre-trained StyleGAN ($\operatorname{D}_{\theta}$) to generate the age-transformed face ($y_\text{tgt}$).
SAM is trained on the FFHQ dataset~\cite{karras_analyzing_2020}, where the training procedure involves producing an age-transformed output $y_{\text{tgt}} = \operatorname{D}_{\theta}(\operatorname{E}_{\theta}(x, a_{\text{tgt}}))$ in a forward pass. This process is supervised by the loss \(\mathcal{L}_{\text{forward}}\), encouraging the re-aged image to be similar to the input image:


\begin{align}
    \mathcal{L}_{\text{forward}}(y_{\text{tgt}}) = &\lambda_{l2} \mathcal{L}_2(y_{\text{tgt}}) +
    \lambda_{lpips} \mathcal{L}_{\text{LPIPS}}(y_{\text{tgt}}) 
    \nonumber \\
    &+ \lambda_{id} \mathcal{L}_{\text{ID}}(y_{\text{tgt}}) +
    {} \lambda_{age} \mathcal{L}_{\text{age}}(y_{\text{tgt}})
\end{align}
\(\mathcal{L}_2\), \(\mathcal{L}_{lpips}\) and \(\mathcal{L}_{\text{ID}}\) matches age-transformed image ($y_\text{tgt}$) to input image ($x$) in pixel space, LPIPS feature space~\cite{lpips} and arcface~\cite{deng_arcface_2022} feature space respectively. $\mathcal{L}_{\mathrm{age}}$ matches the predicted age by a face age detector, $DEX(\cdot)$~\cite{rothe_dex_2015}, with target age \(a_\text{tgt}\): 
\begin{equation}
\mathcal{L}_{\mathrm{age}}\left(y_\text{tgt}\right)= ||a_\text{tgt}-{DEX}\left(y_\text{tgt}\right)||_2
\end{equation}
After the forward pass, SAM encourages the transformed image ($y_{\text{tgt}}$) to be re-transformed back to the input image. This process helps ensure cycle consistency~\cite{cyclegan} and can be formally described as $y_{\text{cycle}} = \operatorname{D}_{\theta}(\operatorname{E}_{\theta}(y_{\text{tgt}}, a_x))$ with the same loss: $\mathcal{L}_{\text{cycle}} (y_{\text{cycle}}) = \mathcal{L}_{\text{forward}} (y_{\text{cycle}})$.
The complete training loss is then given by:
\begin{equation}
    \mathcal{L}_{\text{sam}} = \mathcal{L}_{\text{forward}} (y_{\text{tgt}})
                            + \mathcal{L}_{\text{cycle}} (y_{\text{cycle}})
\end{equation}

\subsection{MyTM: Designing a Personalized Age Adapter}
\label{sec:design}


Training a personalized aging prior from scratch is suboptimal due to the limited availability of personal aging data. To address this, we introduce MyTM, which personalizes a pre-trained age encoder by combining two components: 1) SAM, a pre-trained age encoder that captures a shared global aging prior learned from a diverse set of identities, and 2) Age Adapter, a personalized age adapter network trained exclusively on an individual.
Specifically, we assume the individual has a personal photo collection of $N$ RGB images, \(x_i\), each with an associated ground truth age, \(a_i\), represented as $\mathcal{D} = \left\{(x_i, a_i)\right\}_{i=1}^{N}$.

Our key idea is to update the age-transformed latent code $\mathcal{W}_\text{tgt}^+$ produced by the global age transformation network, SAM, using a personalized adapter network, $AN(\cdot)$. The adapter takes the latent vector $\mathcal{W}_\text{tgt}^+$ predicted by SAM and computes an offset latent vector $\Delta \mathcal{W}_\text{tgt}^+ = \operatorname{AN}(\operatorname{E}_{\theta}(x, a_\text{tgt}), a_\text{tgt})$. As shown in Fig.~\ref{fig:architecture}, we then combine this personalized latent adaptation $\Delta \mathcal{W}_\text{tgt}^+$ with the global latent code $\mathcal{W}_\text{tgt}^+$ and pass the result through the pre-trained StyleGAN2 decoder. Formally,
\setlength{\abovedisplayskip}{3pt}
\setlength{\belowdisplayskip}{3pt}
\begin{align}
 y^{p}_\text{tgt}
 =\operatorname{D}_{\theta}(\eqnmarkbox[darkgray]{global}{\operatorname{E}_{\theta}(x, a_\text{tgt})} + \eqnmarkbox[red]{personal}{\operatorname{AN}(\operatorname{E}_{\theta}(x, a_\text{tgt}), a_\text{tgt})})
 \label{eq:global_personal}
\end{align}
\annotate[yshift=-0em]{below,right}{global}{global aging}
\annotate[yshift=-0em]{below,right}{personal}{personal aging}
\vspace{0.5\baselineskip}

\noindent
By doing so, MyTM enables the integration of partially observed, personalized aging information of an individual—using only images of that person—into the global aging trajectory. Our adapter is based on an MLP architecture, with detailed implementation of the age adapter network available in the supplementary material.

\subsection{MyTM: Loss Funtions}
\label{sec:loss}
Our adapter is trained on a personal photo collection, which is noted as $\mathcal{D}=\left\{(x_{i}, a_{i})\right\}_{i=1}^{N}$. We introduce three loss functions—personalized aging loss, extrapolation regularization, and adaptive w-norm regularization—to integrate global priors with personal data. Additionally, we also use the loss function $\mathcal{L}_{\text{sam}}$ from Sec.~\ref{sec:prelim}, based on SAM, to mitigate the forgetting of global priors~\cite{kirkpatrick_overcoming_2017, nitzan_domain_2023}. 



\subsubsection{Personalized Aging Loss}
After examining the loss formulation of SAM~\cite{alaluf_only_2021}, we notice that the primary source of aging information is the aging loss \(\mathcal{L}_{\text{age}}\), which relies on a pre-trained age classifier. The problem with this global aging loss is that the age classifier is not robust across different ethnicities, styles, and individual aging patterns~\cite{lin_fp-age_2022}. It is often impossible to train a robust aging detector that works well for every individual. Rather than relying on the power of a global age classifier, we propose a personalized aging loss that encourages facial features of the transformed face to be similar to reference images in a similar age range in the training dataset. This encourages the re-aged image to closely resemble how the person looked at that age, ignoring the pose, lighting, and style variations.


We denote the minimum and maximum age of the training dataset $\mathcal{D}=\left\{(x_{i}, a_{i})\right\}_{i=1}^{N}$ as \(a_{min} = \operatorname{min}({a_1 ... a_n})\) and \(a_{max} = \operatorname{max}({a_1 ... a_n})\). During training, we randomly sample a target age \(a_\text{tgt}\) between minimum and maximum age \(a_\text{tgt} \sim \mathcal{U}(a_{min}, a_{max})\).
We create a reference set, \(\mathcal{D}_\text{tgt} = \left\{(x_{j}, a_{j})\right\}_{j=1}^{M}\), which contains actual images of the individual near the target age (\(a_\text{tgt} \pm 3\)-years). We then employ a facial recognition network, arcface \cite{deng_arcface_2022}, to extract identity features and compute the similarity between the age-transformed image $y_\text{tgt}$ and all images in the reference set $\mathcal{D}_\text{tgt}$, and only consider the maximum similarity. Considering maximum similarity over the reference set ensures that the identity recognition networks are not significantly influenced by stylistic differences between images. Formally, we define the personalized aging loss:
\begin{equation}
    \mathcal{L}_{\text{pers-age}} = 1-max\left\{ \left \langle R ( y^{p}_{\text{tgt}} ) , R ( x_{j}) \right \rangle \right\}_{j=1}^{M}
    \label{eq:pers_aging_loss}
\end{equation}
\(R(\cdot)\) is a pretrained arcface~\cite{deng_arcface_2022} network for facial feature recognition, $\left<\cdot,\cdot\right>$ computes the cosine similarity between its argument~\cite{patashnik_styleclip_2021}, and $M$ is the number of images in a reference set $\mathcal{D}_\text{tgt}$ with faces near the target age ($a_\text{tgt} \pm 3$-years).

\subsubsection{Extrapolation Regularization}
When training the adapter network with personalized age loss, we observe that the network's performance can degrade when \(a_\text{tgt}\) falls outside the training age range \([a_{min}, a_{max}]\). Specifically, this degradation manifests as the generated images continuing to resemble the appearance at the boundaries of the training age range (\(a_{min}, a_{max}\)), rather than appropriately aging or de-aging. For instance, as illustrated in row 4 of Fig.~\ref{fig:age_reg} (FADING + Dreambooth), when the training set covers faces aged 30 to 70, the model may overfit, generating faces that still resemble a 30-year-old when \(a_\text{tgt} = 10\). 


To prevent this extrapolation failure, we enforce the preservation of the pre-trained SAM's output during extrapolation. We apply experience replay \cite{nitzan_domain_2023, ruiz_dreambooth_2023}, which encourages the output of our personalized age encoder ($y^{p}_\text{tgt}$) to be similar to that produced by the pre-trained SAM ($y_\text{tgt}$):

\begin{align}
    \mathcal{L}_{\text{reg-extra}} =  \lambda_{l2} \mathcal{L}_2(y^{p}_\text{tgt}, y_\text{tgt}) + \lambda_{\text{LPIPS}} \mathcal{L}_\text{LPIPS}(y^{p}_\text{tgt}, y_\text{tgt}) \nonumber \\
    + \lambda_{\text{ID}} \mathcal{L}_\text{ID}(y^{p}_\text{tgt}, y_\text{tgt})
    \label{eq:reg_extrapolation}
\end{align}



\subsubsection{Adaptive w-norm regularization}
During personalization, we observed that SAM struggles to capture distinct facial feature changes across ages, as illustrated in row 2 of Fig.~\ref{fig:age_reg}. We attribute this issue to the inversion-editability trade-off~\cite{tov_designing_2021, roich_pivotal_2021}. Specifically, the latent codes predicted by SAM are distant from the training distribution and the center of the latent space, $\overline{\mathcal{W}}$, reducing their editing capacity and making personalization challenging.
This trade-off is particularly relevant in facial aging tasks. When the target age is close to the input age, we encourage the latent codes close to SAM’s pre-trained output, preserving inversion accuracy while staying distant from the average latent code $\overline{\mathcal{W}}$. As the target age diverges from the input age, greater deformations in face structures are needed, requiring latent codes nearer to $\overline{\mathcal{W}}$ to facilitate editing.
To address this, we propose adaptive W-norm regularization inspired by~\cite{richardson_encoding_2021}, where \(\mathcal{L}_{\text{reg}} = \lambda_\text{reg} \| \mathcal{W}_\text{tgt}^+ - \overline{\mathcal{W}} \| \) is used to constrain the latent codes.
We further enhance this by making $\lambda_\text{reg}$ a cosine function, $\left<\cdot,\cdot\right>$, of the difference between input and target age \(\Delta_\text{age} = | a_i - a_\text{tgt} | \):
\begin{align}
     \mathcal{L}_{\text{reg}} = \lambda_\text{reg}(\Delta_\text{age}) \| \mathcal{W}_\text{tgt}^+ - \overline{\mathcal{W}} \| \nonumber \\
    \lambda_\text{reg}(\Delta_\text{age}) = 1 - \langle \pi \cdot {\Delta_\text{age}}/{100} \rangle
    \label{eq:reg_w_norm}
\end{align}

\begin{figure*}[!t]
    \centering
    \includegraphics[width=\linewidth]{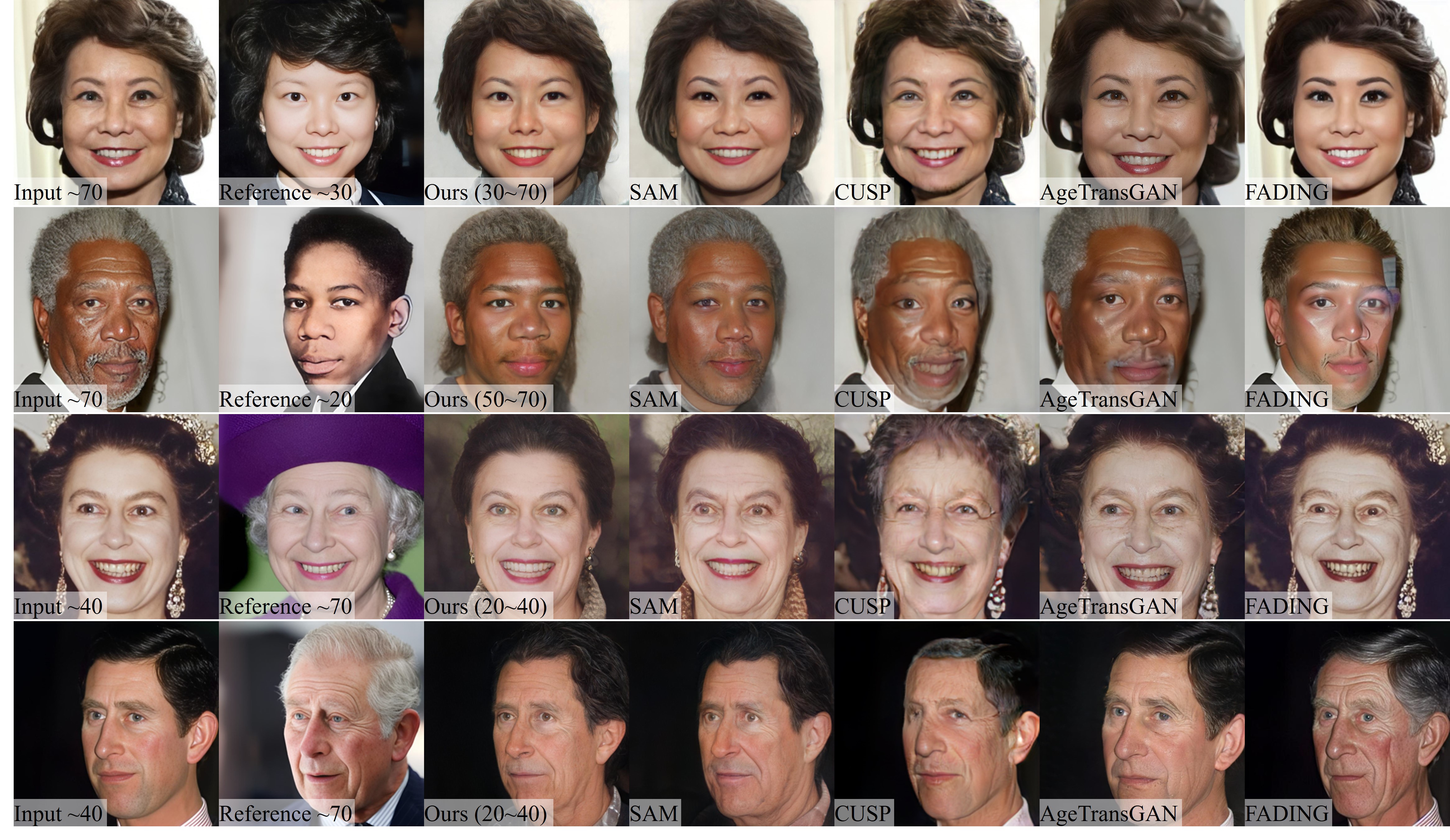}
    \vspace{-2.75em}
    \caption{
    Performance of age transformation techniques for age regression (first two rows) and age progression (last two rows). The first column shows the input image, and the second column provides a reference image of the same person at the target age. MyTM (Ours) is compared against other state-of-the-art methods including SAM~\cite{alaluf_only_2021}, CUSP~\cite{gomez-trenado_custom_2022}, AgeTransGAN~\cite{avidan_agetransgan_2022}, and FADING~\cite{Chen_2023_BMVC}.
    }
    \label{fig:age_compare_pretrained}
    \vspace{-1em}
\end{figure*}

\section{Experiments}
In Sec.~\ref{sec:setup} we first introduce our experimental setup, including datasets, experimental framework, state-of-the-art aging algorithms, and evaluation metrics. In Sec.~\ref{sec:compare} we present comparisons with state-of-the-art baselines, followed by the application of MyTM for video re-aging in Sec.~\ref{sec:video_reaging}. Finally, we present ablation studies in \ref{sec:abla}.

\subsection{Experimental Setup}
\label{sec:setup}

\subsubsection{Dataset}
We curated a dataset of images featuring 12 celebrities spanning a wide age range, including 7 males and 5 females from diverse ethnic backgrounds such as Caucasian, African American, Hispanic, and Asian. For further details, please refer to Table~\ref{tab:dataset} in the supplementary material. 
For each celebrity, we train MyTM on 50 images and evaluate its performance using test images of the same individual at either 40 or 70 years old, depending on the task defined in the experimental setup.

\subsubsection{Experimental Framework}
We consider the following two real-world scenarios where age transformation techniques are heavily used and demand high quality.
\textbf{1) Age regression or de-aging} renders images of an individual to go back in time and is heavily used in VFX for movies and TV shows~\cite{travis_irishman_2019}. Motivated by this, we design an experimental setup where we personalize our aging model by training on images from either a 30$\sim$70 or 50$\sim$70 age range and then evaluate de-aging performance on an unseen image at 70 years old to a target age ($a_\text{tgt} \leq 70$ years old). We sample the target age every 10 years where $a_\text{tgt} \in \{0, 10, 20, 30, 40, 50, 60, 70\}$.
\textbf{2) Age progression or aging} renders images of an individual going forward in time and is used for forensic investigations, missing person searches, or as an emotional support tool to visualize departed loved ones. We design an experimental framework where we personalize our aging model by training on images from 20$\sim$40 years old and evaluate on unseen faces at 40 years old to generate a target age ($a_\text{tgt} \geq 40$), where $a_\text{tgt} \in \{40, 50, 60, 70, 80, 90, 100\}$.


\subsubsection{State-of-the-art Aging Algorithms}
We compare our results with the following state-of-the-art aging methods: (i) SAM \cite{alaluf_only_2021}, which uses a pre-trained StyleGAN2 decoder and trains an age encoder on FFHQ~\cite{karras_style-based_2019}. It treats aging as a continuous process, enabling fine-grained control over transformations. (ii) CUSP \cite{gomez-trenado_custom_2022}, which jointly trains both an age encoder and decoder on FFHQ-Aging~\cite{or-el_lifespan_2020}. While effective for age transformations, it lacks fine-grained control due to its reliance on predefined age group-based transformations, limiting editing capabilities and supporting resolutions only up to $256\times256$. (iii) AgeTransGAN \cite{avidan_agetransgan_2022}, an encoder-decoder architecture, also limited by age group-based transformations, similar to CUSP. (iv) FADING~\cite{Chen_2023_BMVC}, which inverts images into the latent space of a face diffusion model using NTI~\cite{mokady_null-text_2022}, then edits them through p2p~\cite{hertz_prompt--prompt_2022}.
We also introduce additional personalization baselines: (v) SAM Pers. f.t., naively fine-tune SAM on personal images; (vi) FADING + Dreambooth, personalizes FADING by following the Dreambooth approach~\cite{ruiz_dreambooth_2023} with the prompt \textit{"photo of a [$a_\text{tgt}$] year old [sks] man/woman"}.

\subsubsection{Evaluation Metrics}
Following the evaluation protocols in prior aging baselines \cite{alaluf_only_2021, gomez-trenado_custom_2022, avidan_agetransgan_2022, Chen_2023_BMVC}, we evaluate our personalized age transformation results in terms of age accuracy and identity preservation, using the following metrics to evaluate the re-aged results quantitatively:

\noindent \textit{$\bullet$ Age Accuracy ($\text{Age}_{MAE}$).} Following~ \cite{alaluf_only_2021, gomez-trenado_custom_2022},
we define age mean absolute error as $\text{Age}_{MAE} = | \hat{a_\text{tgt}} - a_\text{tgt} | $, where $\hat{a_\text{tgt}}$ is predicted by FP-Age~\cite{lin_fp-age_2022}.

\noindent \textit{$\bullet$ Identity Preservation ($\text{ID}_{sim}$).}
Previous works~\cite{zoss_production-ready_2022, teng_exploring_2023} evaluate identity preservation by comparing the re-aged face to the input face. 
    However, facial recognition systems, such as arcface~\cite{deng_arcface_2022}, demonstrate a strong dependence on age and thus favor age consistency between the re-aged face and the input face~\cite{alaluf_only_2021}, favoring algorithms that perform small changes. We address this problem by creating reference image sets of the individual near the target age which are not used in training, and then calculate the identity similarity to the reference images at the target age, in contrast to using the input image.
    Formally,

    \begin{equation}
    \text{ID}_{sim} \left (y_{ \text{tgt}}\right ) = max\left\{ \left \langle R ( y_{ \text{tgt}} ) , R ( x_{j}) \right \rangle \right\}_{j=1}^{M}
    \label{eq:id_metric}
    \end{equation}
    where \(R(\cdot)\) is a pretrained arcface~\cite{deng_arcface_2022} network for facial feature recognition and $x_{j}$ belongs in reference image set near the target age ($a_\text{tgt} \pm 3$-years). We report the average $\text{ID}_{sim}$ across all sampled target ages.

\begin{table}[t]
    \centering
    \caption{
    Performance of age regression where a 70-year-old face is de-aged to a target age $a_{tgt} \leq 70$. We also evaluate MyTM (Ours) using 20-year ($a_\text{tgt} \in 50\sim70$) and 40-year ($a_\text{tgt} \in 30\sim70$) age ranges in the training data. \textbf{Bold} indicates the best results, while \underline{underlined} denotes the second-best. 
    }
    \vspace{-1em}
    \resizebox{\columnwidth}{!}{%
    \begin{tabular}{lcccc}
    \toprule 
    \multirow{2.5}{*}{Method} & \multirow{2.5}{*}{$\text{Age}_{MAE} (\downarrow$)} & \multicolumn{3}{c}{$\text{ID}_{sim} (\uparrow$)} \\
    \cmidrule(lr){3-5}
    & 
    & $a_\text{tgt} \leq 70$ & $a_\text{tgt} \in 50\sim 70$ & $a_\text{tgt} \in 30\sim 70$ \\
    \midrule
    SAM~\cite{alaluf_only_2021} & 8.1 & 0.49 & 0.58 & 0.53 \\
    + Pers. f.t. (50$\sim$70) & 8.2 & 0.48 & 0.58 & - \\
    + Pers. f.t. (30$\sim$70) & 9.2 & 0.49 & - & 0.53 \\
    CUSP~\cite{gomez-trenado_custom_2022} & 11.0 & 0.39 & 0.44 & 0.42 \\
    AgeTransGAN~\cite{avidan_agetransgan_2022} & 11.1 & 0.53 & 0.65 & 0.58 \\
    FADING~\cite{Chen_2023_BMVC} & 8.9 & 0.60 & 0.72 & 0.66 \\
    \makecell[l]{+ Dreambooth~\cite{ruiz_dreambooth_2023} (50$\sim$70)} & 25.9 & 0.63 & \tabfirst{0.78} & - \\
    \makecell[l]{+ Dreambooth~\cite{ruiz_dreambooth_2023} (30$\sim$70)} & 23.0 & 0.64 & - & \tabsecond{0.70} \\
    Ours (50$\sim$70) & \tabfirst{7.7} & \tabsecond{0.65} & \tabsecond{0.76} & - \\
    Ours (30$\sim$70) & \tabsecond{7.8} & \tabfirst{0.67} & - & \tabfirst{0.72} \\
    \bottomrule
    \end{tabular}%
    }

    \label{tab:age_reg}
    \vspace{-1.5em}
\end{table}

\subsection{Comparison with Age Transformation Methods}
\label{sec:compare}
\subsubsection{Age Regression}
We use two age ranges of personal photos—40 years (ages 30$\sim$70) and 20 years (ages 50$\sim$70)—to examine the impact of training age span. 
Results are presented in Table~\ref{tab:age_reg}, with visual examples of pre-trained methods shown in Fig.~\ref{fig:age_compare_pretrained}. For a detailed visual comparison across all ages (0$\sim$100), please refer to Fig.~\ref{fig:age_baseline}. Additionally, visual results are compared against personalized methods in Fig.~\ref{fig:age_reg}.

Compared to pre-trained baselines (SAM, CUSP, AgeTransGAN, and FADING), our method achieves superior identity preservation ($\text{ID}_{sim}$), with an 11.7\% improvement (0.67 vs. 0.60) in $\text{ID}_{sim}$ over the best-performing method, FADING.
This improvement is also maintained during interpolation (e.g., when $a_\text{tgt} \in 30\sim70$), producing a 9.0\% increase in $\text{ID}_{sim}$ (0.72 vs. 0.66) compared to FADING, even when FADING overfits to the input image via NTI~\cite{mokady_null-text_2022}, favoring its $\text{ID}_{sim}$ score for smaller age gaps. 

Compared to other personalized methods, our approach achieves both high age accuracy ($\text{Age}_{MAE}$) and strong identity preservation. SAM + Pers. f.t. shows minimal improvement over SAM alone, underscoring the effectiveness of our proposed loss function in Sec.~\ref{sec:loss}. While FADING + Dreambooth~\cite{ruiz_dreambooth_2023} (50$\sim$70) records a slight improvement over ours in $\text{ID}_{sim}$ (0.78 vs. 0.76), it fails to maintain age accuracy ($\text{Age}_{MAE}$ 25.9 vs. 7.7) and overfits to the training age range, limiting its ability to generalize to unseen ages.

\subsubsection{Age progression}
We perform age progression with a 20-year (ages $20\sim40$) range of personal photos.
Age progression specifically evaluates the extrapolation ability of our technique to ages not seen in training. Quantitative results are presented in Table~\ref{tab:age_prog}, with visual comparisons to pre-trained methods shown in Fig.~\ref{fig:age_compare_pretrained}, and a full visual comparison provided in  Fig.~\ref{fig:age_baseline} and Fig.~\ref{fig:age_prog}.

Our model outperforms pre-trained baselines, achieving the highest age accuracy (6.3) and best identity preservation (0.70 for $a_\text{tgt} \geq 40$ and 0.78 for $a_\text{tgt} \in 40 \sim 60$), due to the benefits of personalization. 
As shown in Fig.~\ref{fig:age_baseline}, FADING often produces poor results when the target age differs greatly from the input age, due to NTI + p2p editing~\cite{rout_semantic_2024}.
Compared to other personalized methods, FADING + Dreambooth achieves slightly better $\text{ID}_{sim}$ than our model (0.72 vs. 0.70). However, it struggles to extrapolate to unseen ages, resulting in a high $\text{Age}_{MAE}$ of 20.2.


\begin{table}[t]
    \centering
    \caption{
    Performance of age progression where a 40-year-old face is aged to a target age $a_{tgt} \geq 40$. 
    We evaluate MyTM (Ours) using 20-year ($a_\text{tgt} \in 40\sim60$) and $a_\text{tgt} \geq 40$ age ranges in the training data. \textbf{Bold} indicates the best results, while \underline{underlined} denotes the second-best. Note that FADING + Dreambooth has the lowest aging accuracy, as measured by $\text{Age}_{MAE}$.
    }
    \vspace{-0.5em}
    \resizebox{\columnwidth}{!}{%
    \begin{tabular}{lccc}
    \toprule 
    \multirow{2.5}{*}{Method} & \multirow{2.5}{*}{$\text{Age}_{MAE} (\downarrow$)} & \multicolumn{2}{c}{$\text{ID}_{sim} (\uparrow$)} \\
    \cmidrule(lr){3-4}
    &
    & $a_\text{tgt} \geq 40$ & $a_\text{tgt} \in 40 \sim 60$ \\
    \midrule
    SAM~\cite{alaluf_only_2021} & \tabsecond{6.9} & 0.54 & 0.58 \\
    + Pers. f.t. (20$\sim$40) & 10.3 & 0.56 & 0.59 \\
    CUSP~\cite{gomez-trenado_custom_2022} & 7.3 & 0.44 & 0.48 \\
    AgeTransGAN~\cite{avidan_agetransgan_2022} & 8.5 & 0.61 & 0.65 \\
    FADING~\cite{Chen_2023_BMVC} & 7.6 & 0.62 & 0.71 \\
    \makecell[l]{+ Dreambooth~\cite{ruiz_dreambooth_2023} (20$\sim$40)} & 20.2 & \tabfirst{0.72} & \tabsecond{0.77} \\
    Ours (20$\sim$40) & \tabfirst{6.3} & \tabsecond{0.70} & \tabfirst{0.78} \\
    \bottomrule
    \end{tabular}%
    }
    \label{tab:age_prog}
\end{table}

\begin{figure}[t]
    \centering    \includegraphics[width=\columnwidth]{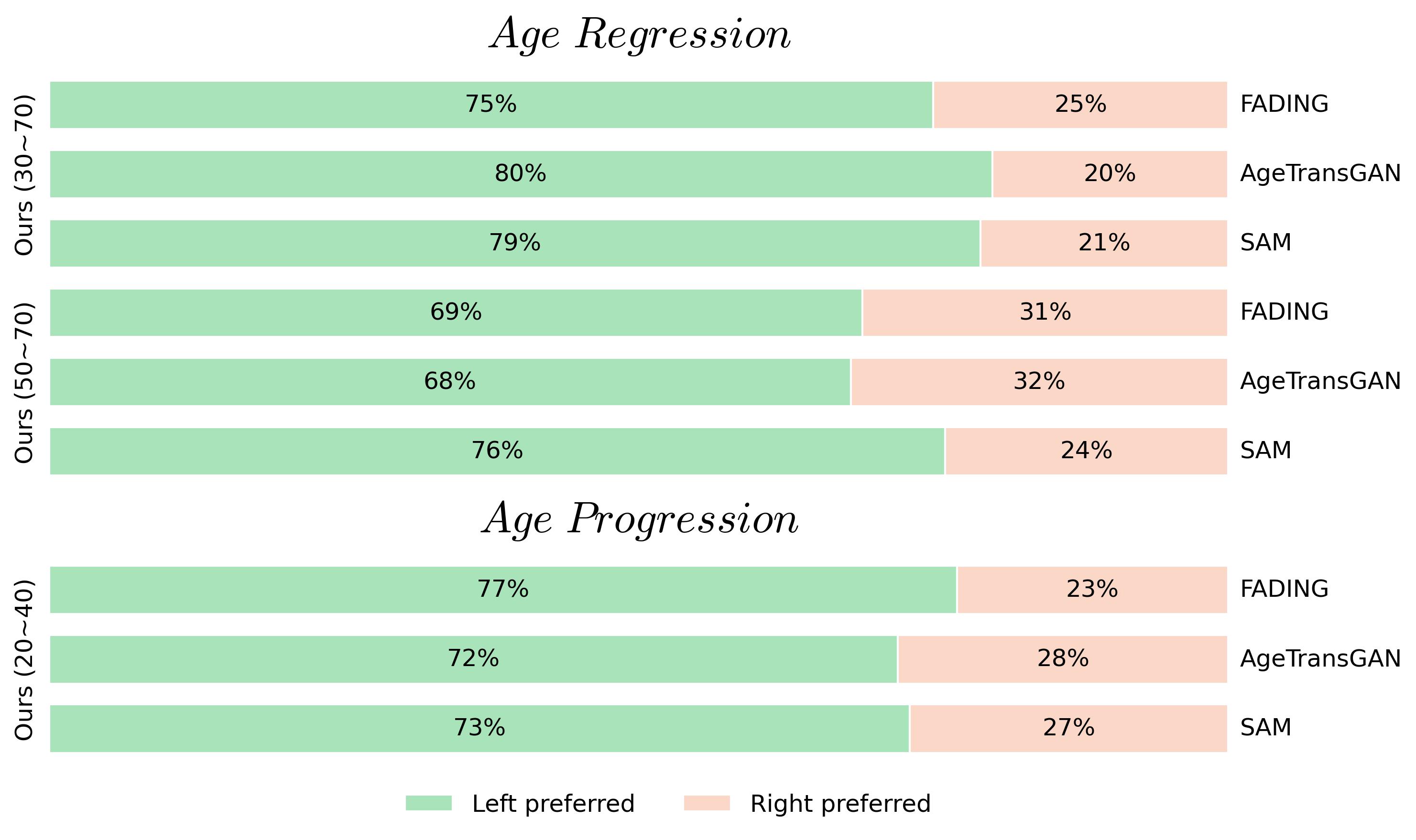}
    \vspace{-2.5em}
    \caption{User study comparing our method with baselines—FADING, AgeTransGAN, and SAM—for age regression ($a_\text{tgt} \leq 70$) and age progression ($a_\text{tgt} \geq 40$). 
    We present the percentage of user preference for our method over the baselines.}
    \label{fig:user_study}
    \vspace{-1.5em}
\end{figure}

\subsubsection{User Studies}
We conduct user studies to qualitatively evaluate our method through pairwise human evaluations. In each pair, users see the original input image alongside two re-aged results at the target age $a_\text{tgt}$—one generated by an existing method and the other by ours, presented in random order. Users also receive reference images showing the person’s face near $a_\text{tgt}$ and are asked to select the result that best matches the reference images while preserving the style of the original input image.

    We evaluate our method across two age regression tasks (30$\sim$70 and 50$\sim$70) and one age progression task (20$\sim$40), totaling three tasks. 
    We then sample one input and re-aged image pair per celebrity, resulting in 10 pairs for each age regression task and 8 pairs for the age progression task. For each pairwise comparison, we collected 24 responses for FADING, 25 for AgeTransGAN, and 29 for SAM.
    As shown in Fig.~\ref{fig:user_study}, our method is significantly preferred over the baselines across all re-aging tasks.

\subsection{MyTM for Video Re-aging}
\label{sec:video_reaging}
\begin{figure}[t]
    \centering
    \includegraphics[width=\columnwidth]{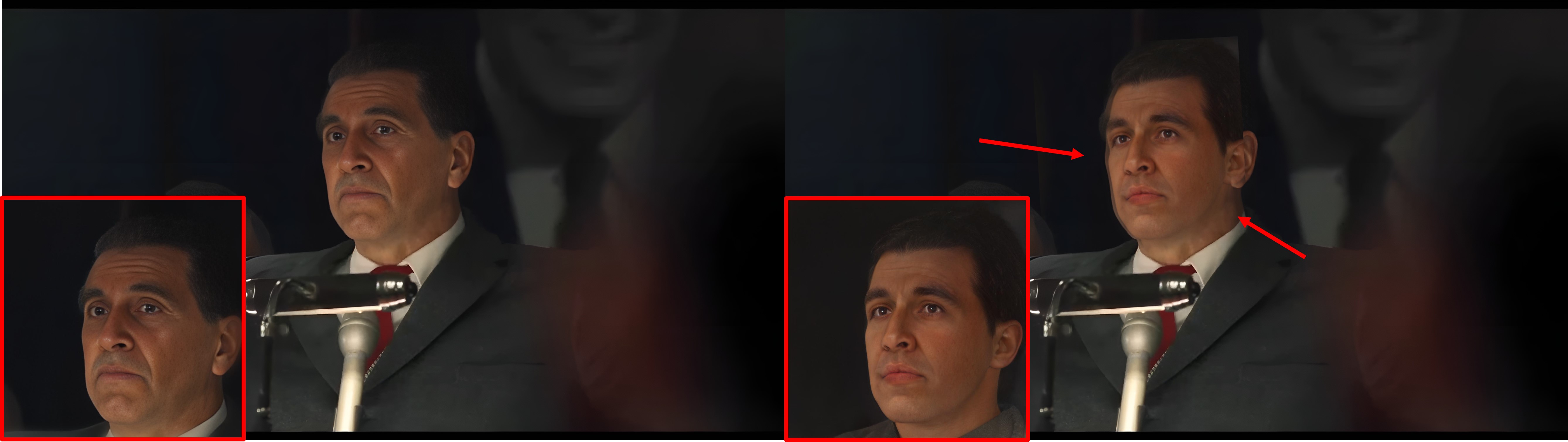}
    \vspace{-2em}
    \caption{
    Naive pasting of MyTM's re-aged face onto a video frame of \textit{Al Pacino} from \textit{The Irishman}. Left: The original frame, with the aligned face shown in the bottom-left corner. Right: The re-aged face generated by MyTM is naively pasted back onto the original frame.
    Red arrows highlight visible artifacts. Zooming in is recommended for a clearer view.
    }
    \label{fig:naive_paste_back}
    \vspace{-1em}
\end{figure}

\noindent
Having established a 2D personalized aging prior, we now extend our approach to video-based face re-aging. A straightforward method would involve applying MyTM to re-age each frame individually and then pasting the transformed face back onto the original frame. However, this naive approach is prone to two key issues.
(a) Face Detection Errors: Face detection algorithms such as dlib~\cite{dlib09} may fail to detect faces in challenging frames or may produce inaccurate landmark estimates. These inconsistencies lead to misalignment during the warping process, resulting in noticeable artifacts.
(b) Non-Rigid Transformations: The re-aging process inherently alters the facial structure and contours. Because these changes are non-rigid, aligning and pasting the modified face onto the original frame becomes theoretically infeasible without distortion.
An example of this failure is illustrated in Fig.~\ref{fig:naive_paste_back}, where the re-aged face is naively pasted back onto the original frame using landmark-based warping. For clarity, the original frame's head is shown with partial transparency, highlighting the misalignment caused by naive compositing.

\begin{figure}[t]
    \centering
    \includegraphics[width=\columnwidth]{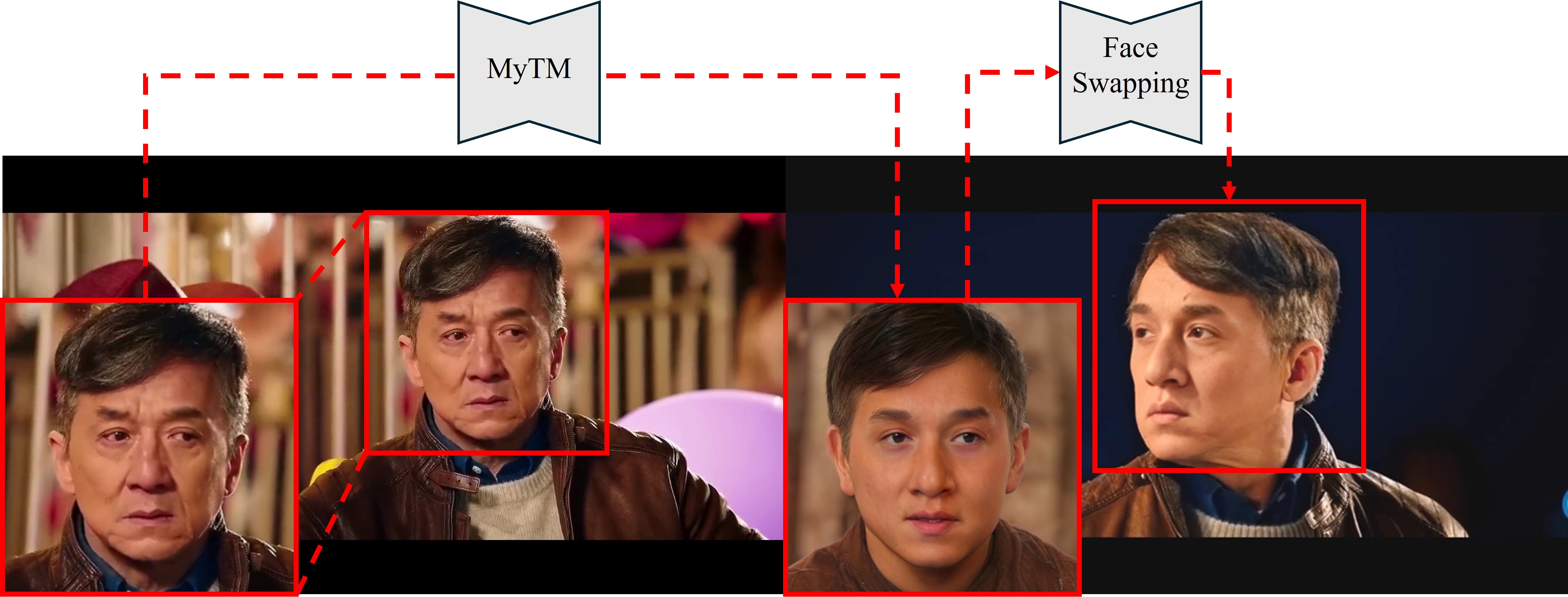}
    \vspace{-2em}
\caption{
Face-swapping for video re-aging on a clip of \textit{Jackie Chan} from \textit{Bleeding Steel}. Left: Keyframe re-aged using MyTM. Right: The re-aged face is transferred to the remaining frames via face-swapping, preserving temporal consistency across the video.
}
    \label{fig:video_reaging}
    \vspace{-1em}
\end{figure}

To this end, we build upon face-swapping techniques by utilizing Inswapper\footnote{https://github.com/deepinsight/insightface}, a widely adopted black-box model~\cite{wang_instantid_2024} for video re-aging.
Given a source video, we manually select a keyframe in a near-frontal pose with minimal occlusion and motion blur as the basis for re-aging. MyTM is then applied to this keyframe to transform the face image to the desired target age, generating a personalized re-aged face. Next, for each video frame, the face in the current frame and the re-aged face are input into the swapping model to generate the final re-aged result. This re-aged face is then pasted back onto the current frame using landmark-based warping.
This framework requires only a single re-aged face for swapping, ensuring strong temporal consistency while preserving personalized facial identity. Our video re-aging pipeline is illustrated in Fig.~\ref{fig:video_reaging}. For further details on temporal consistency, please refer to the supplement.

\subsection{Ablation Study}
\label{sec:abla}

\subsubsection{Effect of Dataset Size}
\begin{figure}[t]
\begin{minipage}{\columnwidth}
    \centering
    \includegraphics[width=\columnwidth]{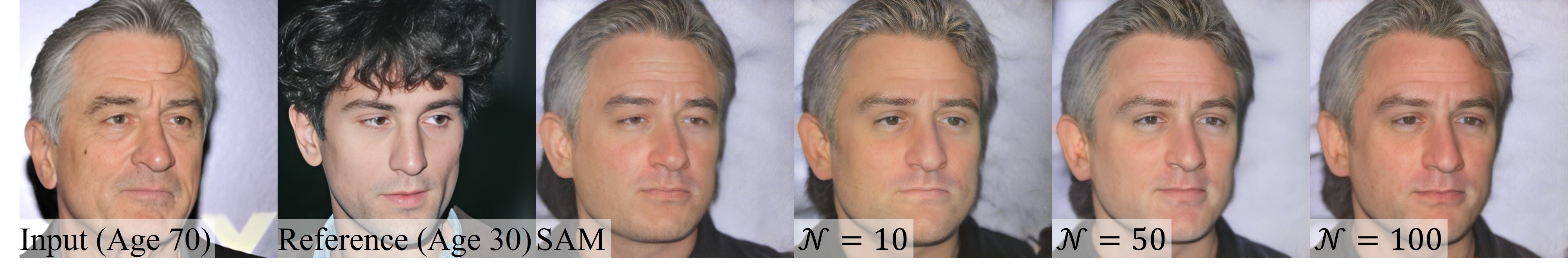}

    \centering
    \resizebox{\columnwidth}{!}{%
        \begin{tabular}{cccccc}
            \toprule
            \multirowcell{3}[-0.4ex]{Dataset $\mathcal{D}$ \\ Size Ablations}
            & Metric & SAM & $\mathcal{N} = 10$ & $\mathcal{N} = 50$ & $\mathcal{N} = 100$  \\
            \cmidrule(lr){2-6} & $\text{Age}_{MAE} (\downarrow$) & 8.1 & 8.5 & 7.8 & 8.0 \\
            &
            ID$_\text{sim} (\uparrow)$
            &
            0.49
            &
            0.58 
            &
            0.67
            &
            0.67
            \\
            \bottomrule
        \end{tabular}%
    }
    \vspace{-1em}
    \caption{ Effect of training dataset size $\mathcal{D}$ on personalization. MyTM is trained on ages 30$\sim$70 and tested for $a_\text{tgt} \leq 70$. Visual examples of \textit{Robert De Niro} are shown at the top, with quantitative results displayed below. MyTM achieves personalized re-aging with as few as 10 images, with 50 images providing optimal performance.}
    \label{fig:ablation_dataset_size}
\end{minipage}
\end{figure}

We investigate the impact of training dataset size on MyTM by sampling subsets of images for each celebrity, with sizes of 10, 50, and 100. We then assess MyTM’s performance on the age regression task (ages 30$\sim$70), which demands the largest training age range.
We report the average $\text{ID}_{sim}$ in Fig.~\ref{fig:ablation_dataset_size}. Results indicate a significant performance improvement from 10 to 50 images, with minimal gains from 50 to 100 images. Consequently, we use 50 images for personalization, unless otherwise noted.

\subsubsection{Effect of MyTM on Non-celebrities}
We collect YouTube videos of five non-celebrities, where individuals documented their age transformation with an average age span of 20 to 54 years. From these videos, we create a dataset of approximately 50 training images per person and train personalized models on ages 20-40, evaluating them using held-out test images at age 40. To assess performance, we measure identity preservation for interpolation (ages 20-40) and extrapolation (>40), as well as transformed age accuracy, as shown in Table~\ref{tab:non-celeb}.
Our results demonstrate that MyTM achieves overall the best identity preservation and age accuracy for both celebrity and in-the-wild non-celebrity data, both quantitatively and qualitatively.

\begin{table}[t]
    \caption{
    Performance of both age regression and progression using non-celebrity subjects, where a 40-year-old face is re-aged to a target age $a_\text{tgt}$. \textbf{Bold} values indicate the best performance.
    }
    \centering
    \resizebox{\columnwidth}{!}{%
    \begin{tabular}{lccc}
    \toprule 
    \multirow{2.5}{*}{Method} & \multirow{2.5}{*}{$\text{Age}_{MAE} (\downarrow$)} & \multicolumn{2}{c}{$\text{ID}_{sim} (\uparrow$)} \\
    \cmidrule(lr){3-4}
    & 
    & $a_\text{tgt} \in 20\sim 40$ & $a_\text{tgt} \geq 40$ \\
    \midrule
    SAM~\cite{alaluf_only_2021} & 12.1 & 0.62 & 0.53 \\
    AgeTransGAN~\cite{avidan_agetransgan_2022} & 10.3 & 0.67 & 0.61 \\
    FADING~\cite{Chen_2023_BMVC} & 13.1 & 0.72 & 0.67 \\
    \makecell[l]{+ Dreambooth~\cite{ruiz_dreambooth_2023} (20$\sim$40)} & 20.9 & 0.78 & 0.69 \\
    Ours (20$\sim$40) & \tabfirst{9.3} & \tabfirst{0.83} & \tabfirst{0.74} \\
    \bottomrule
    \end{tabular}%
    }
    \label{tab:non-celeb}
\end{table}

\subsubsection{Effect of Proposed Loss Functions and Architecture}
\begin{figure}[t]
\begin{minipage}{\columnwidth}
    \centering
    \includegraphics[width=0.95\columnwidth]{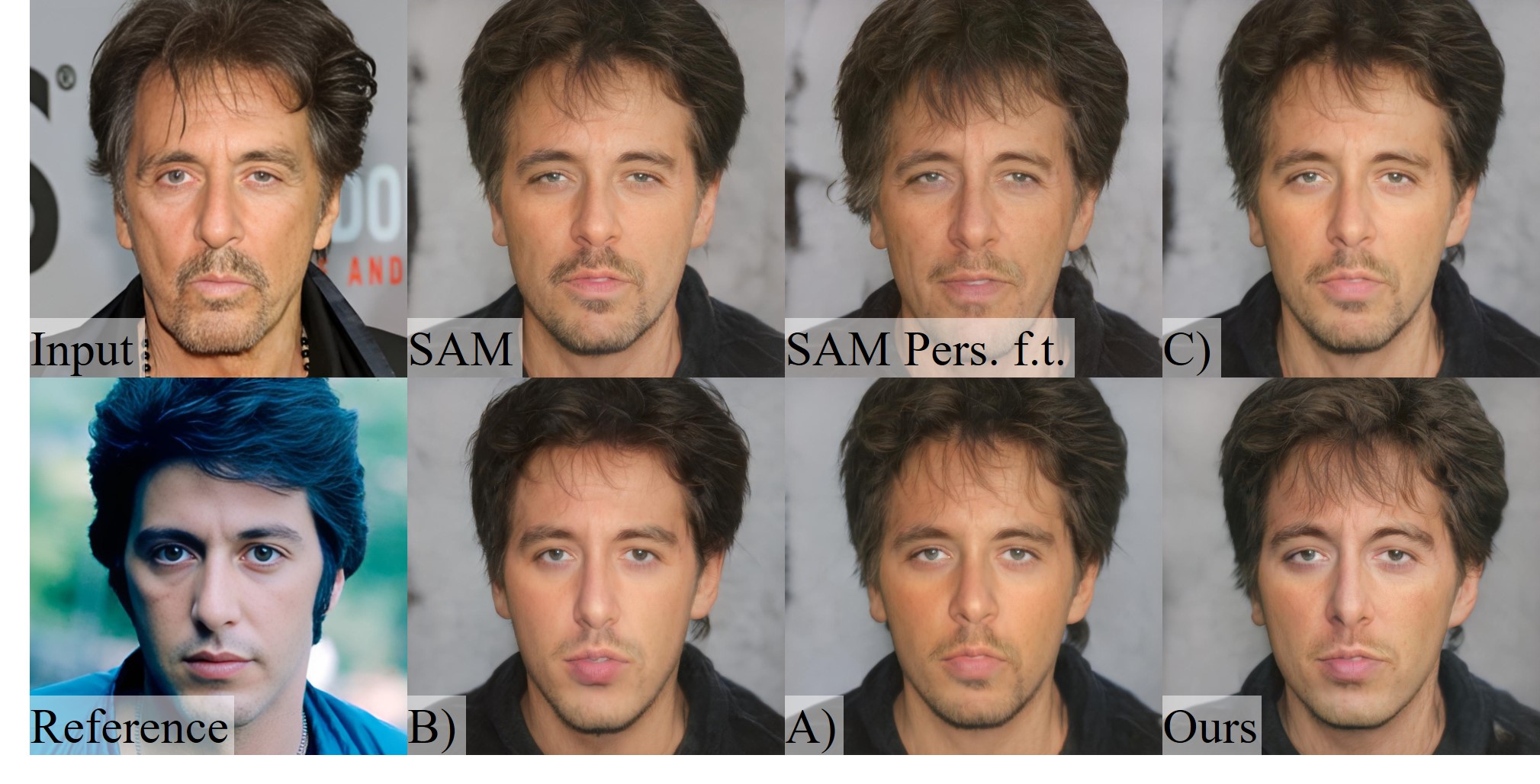}
    \centering
    \resizebox{\columnwidth}{!}{%
    \begin{tabular}{lcccccc}
    \toprule
    \multirow{2}{*}{\makecell[c]{Model}} & \multirow{2}{*}{\makecell[c]{Adapter \\ Network}} & \multirow{2}{*}{\makecell[c]{Extrapolation \\ Reg}} & \multirow{2}{*}{\makecell[c]{Personalized Aging \\ Loss}} & \multirow{2}{*}{\makecell[c]{Adaptive W-norm \\ Reg}} & \multirow{2}{*}{\makecell[c]{ID$_{sim} (\uparrow)$}} & \multirow{2}{*}{\makecell[c]{Age$_{MAE} (\downarrow)$}}
    \\
    \\
    \midrule
    Ours & \checkmark & \checkmark & \checkmark & \checkmark & 0.67 & 10.5 \\
    A) & - & \checkmark & \checkmark & \checkmark & 0.65 & 11.0 \\
    B) & - & - & \checkmark & \checkmark & 0.65 & 17.9 \\
    C) & - & - & - & \checkmark & 0.55 & 12.9 \\
    SAM Pers. f.t. & - & - & - & - & 0.46 & 11.2 \\
    SAM & - & - & - & - & 0.45 & 11.1 \\
    \bottomrule
    \end{tabular}%
    }
    \vspace{-1em}
    \caption{
    Contributions of our proposed loss functions and the adapter network for the age regression task, trained on ages 30$\sim$70 and tested for $a_\text{tgt} \leq 70$ on \textit{Al Pacino}.
    }
    \label{fig:ablation_loss_net}
\end{minipage}
\end{figure}

We analyze the effectiveness of our proposed network architecture and loss functions by conducting an ablation study in Fig.~\ref{fig:ablation_loss_net}. We begin with SAM and progressively introduce each proposed component, including custom loss terms and the adapter network. For the age regression task, we train MyTM on ages spanning 30 to 70, testing with target ages $a_\text{tgt} \leq 70$. Our proposed Personalized Aging Loss yields the most improvement in $\text{ID}_{sim}$. 

Our results in Fig.~\ref{fig:ablation_loss_net} show that incorporating the extrapolation loss significantly improves Age$_\text{MAE}$ (from 17.9 to 11.0) without affecting identity similarity (ID$_\text{sim}$ remains at 0.65). This suggests that a personalized prior alone is insufficient for extrapolation, due to its training on a limited temporal range. To address this, we incorporate a global aging prior by encouraging MyTM’s output to align with that of a pre-trained SAM model. This fusion of personalized and global priors enables more accurate age transformation while preserving identity, further supporting the observed improvement in Age$_\text{MAE}$ with stable ID$_\text{sim}$.


\subsubsection{Direct Video Editing vs. Face-Swapping}
To further validate the effectiveness of our face-swapping pipeline in Sec.~\ref{sec:video_reaging}, we compare it against direct video-based re-aging methods~\cite{10.1145/3550469.3555382, 10.1007/978-3-031-19784-0_21}. Specifically, we benchmark our method against STIT~\cite{10.1145/3550469.3555382} and VideoEditGAN~\cite{xu_temporally_2022} using a set of 14 video clips. To assess visual quality, we conducted a user study using randomly ordered pairwise comparisons. Participants evaluated each pair based on temporal consistency and identity similarity. Responses from 28 participants were collected, and user preferences for our method over the baselines are summarized in Table~\ref{tab:user_study_video}.

Our results indicate that direct video editing methods exhibit lower temporal consistency and weaker identity preservation compared to our face-swapping approach. STIT~\cite{10.1145/3550469.3555382} exhibits noticeable flickering, especially in videos with fast head movements, extreme poses, or challenging lighting conditions. VideoEditGAN~\cite{xu_temporally_2022} demonstrates better temporal consistency than STIT but still lacks temporal consistency and personalized re-aging compared to ours.
Moreover, these methods require per-frame PTI~\cite{roich_pivotal_2021} inversion and optimization, leading to runtimes $>$3 hours per video on a single A6000 GPU. In contrast, our pipeline completes the same task in $<$5 minutes on the same hardware, while achieving better temporal coherence and more personalized re-aging.

\begin{table}[t]
    \caption{
    User study comparing our face-swapping results with direct video re-aging baselines, STIT and VideoEditGAN. We evaluate temporal consistency and identity preservation, reporting user preference percentages (User\%) for ours over each baseline.
    }
    \centering
    \resizebox{\columnwidth}{!}{%
    \begin{tabular}{lcc}
    \toprule
    Method & Temporal Consistency (User\%) & ID (User\%) \\
    \midrule
    Ours vs. STIT~\cite{10.1145/3550469.3555382} & 86\%   &   93\% \\
    Ours vs. VideoEditGAN~\cite{xu_temporally_2022}  & 71\% &  79\% \\
    \bottomrule
    \end{tabular}%
    }

    \label{tab:user_study_video}
\end{table}
\section{Conclusion}

We present MyTimeMachine, a personalized facial age transformation method leveraging individual photo collections and global aging priors, that outperforms existing approaches.

\subsection{Limitations}
While our model performs effective age transformations (see Fig.~\ref{fig:age_regression_oprah}, supp. pdf), it can struggle with accessories such as glasses, largely due to limitations in the e4e encoder~\cite{tov_designing_2021}.
Furthermore, the pre-trained SAM model has difficulty in 1) modifying hair color, particularly transitioning to or from white. For example, hair does not turn white when the target age $\geq$ 80 — a common challenge in other aging works~\cite{li_continuous_2021, tang_face_2018}. Future work could focus on improving global aging models or developing specialized post-editing techniques for hairstyles.
2) producing red-eye artifacts when generating older faces. Although our proposed w-norm regularization mitigates these issues, they are not fully resolved.

\subsection{Ethical Considerations}
Facial aging is a complex and inherently challenging problem, and even with personalization, our model may lack robustness across all underrepresented populations. Our approach also has the potential to produce manipulated images of real individuals, which poses a significant societal risk. 
Although such concerns are common to generative models, we explicitly call for the fair and responsible use of our method. It is intended for research and positive applications only. Misuse that infringes on privacy, spreads misinformation, or causes harm is strongly discouraged. Future work should include safeguards such as synthetic image detection and fairness evaluation.


\begin{acks}
This research was supported in part by Lenovo Research (Morrisville, NC). We gratefully acknowledge the invaluable support and assistance of the members of the Mobile Technology Innovations Lab. This work was also supported in part by the National Science Foundation under Grant No. 2213335.
\end{acks}

\clearpage

\begin{figure*}[th]
    \centering
    \includegraphics[width=\linewidth]{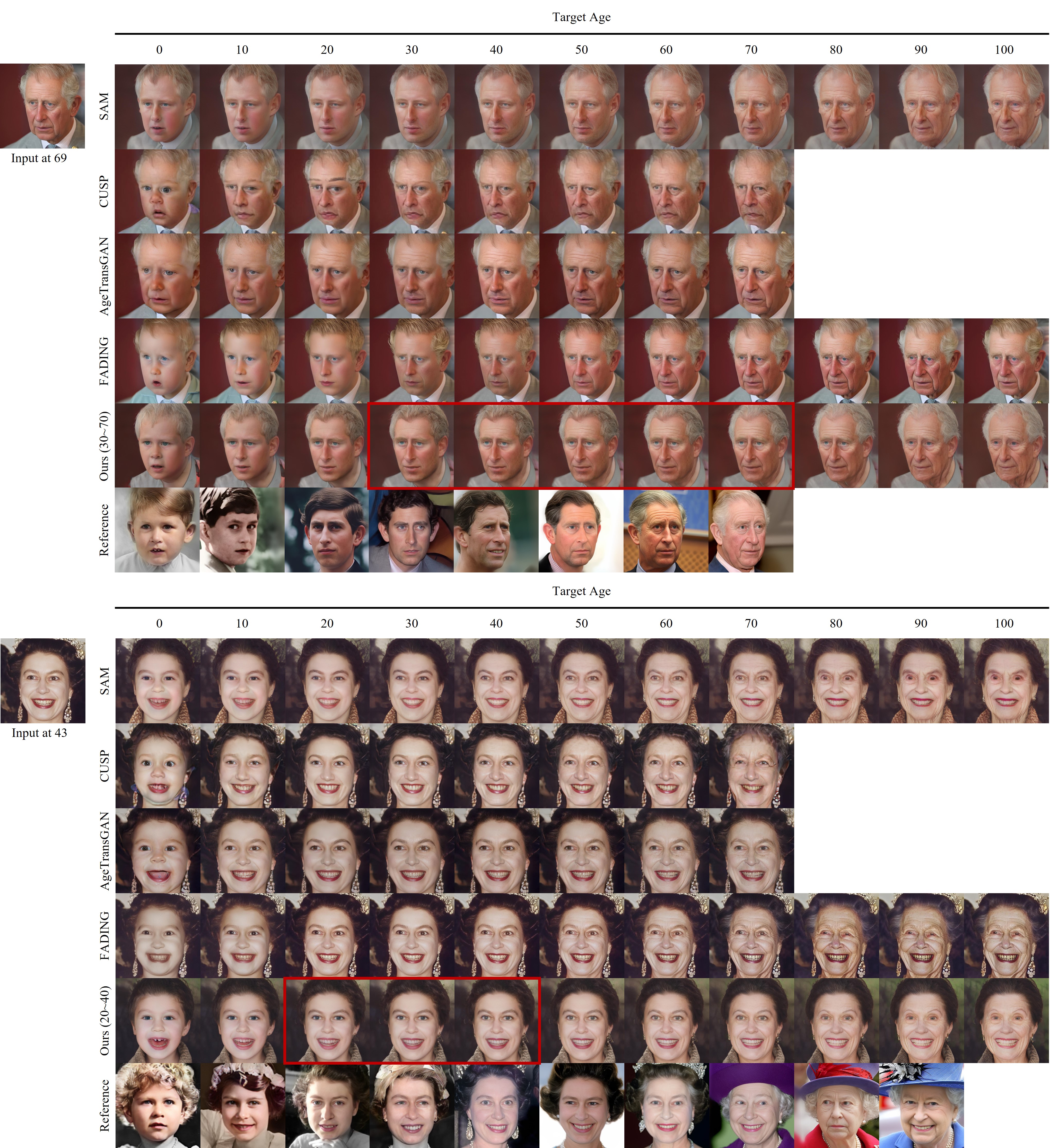}
\caption{
Performance of age transformation techniques for age regression (top) and age progression (bottom). For age regression, MyTM (Ours) is trained across a 40-year range (ages 30 to 70), while for age progression, it is trained over a 20-year range (ages 20 to 40). Personalized training data age ranges are marked in red. A reference image of the same person, taken within 3 years of the target age, is included for comparison.
}
    \label{fig:age_baseline}
\end{figure*}


\begin{figure*}[th]
    \centering
    \includegraphics[width=\linewidth]{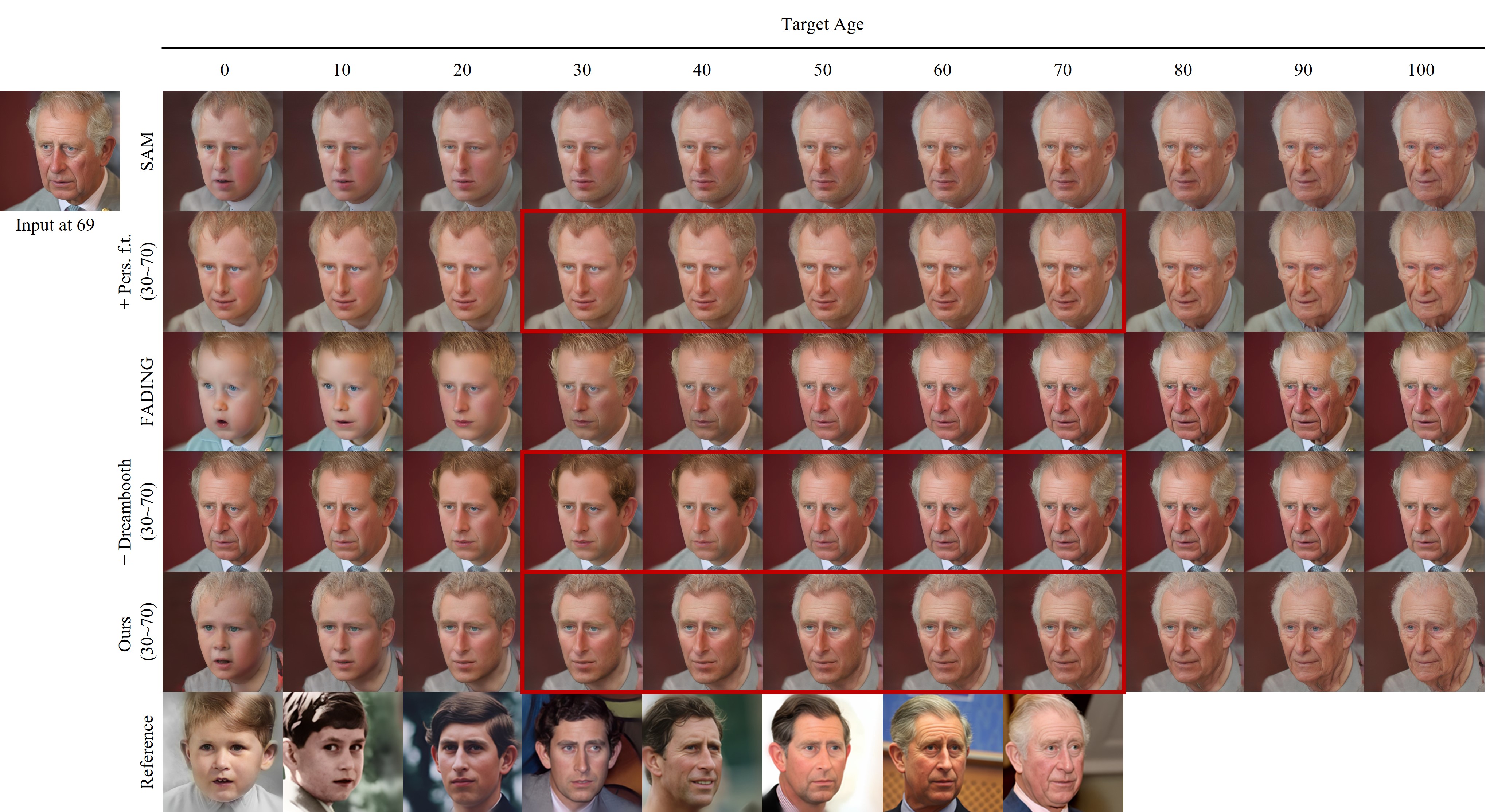}
\end{figure*}

\begin{figure*}[th]
    \vspace{-1em}
    \centering
    \includegraphics[width=\linewidth]{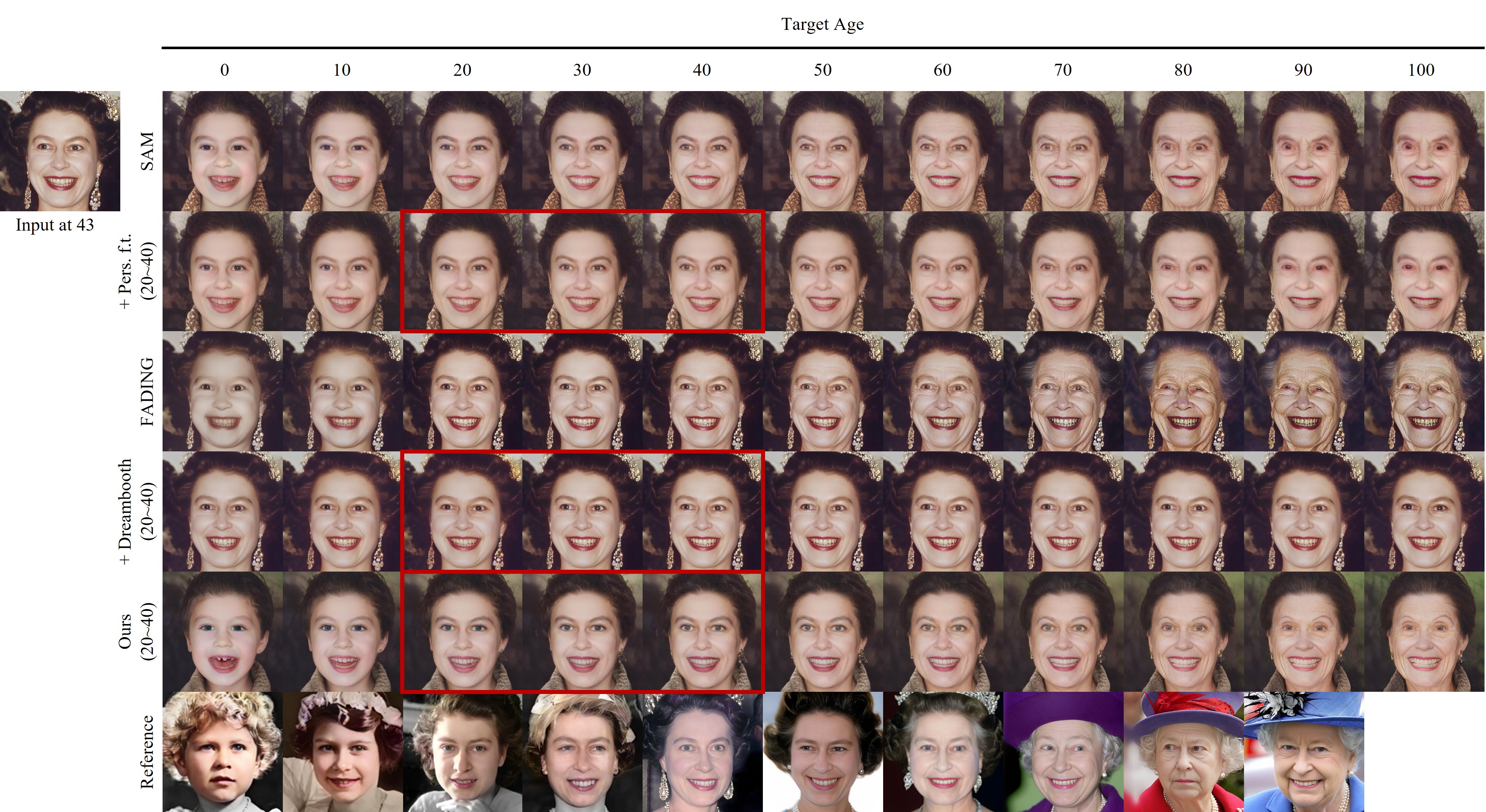}
\caption{Performance of personalized age transformation techniques for age regression (top) and age progression (bottom). The input test images match those in Fig.~\ref{fig:age_reg} (top) and Fig.~\ref{fig:age_prog} (bottom) for consistency. For age regression, MyTM (Ours) is trained across a 40-year range (ages 30 to 70), while for age progression, it is trained over a 20-year range (ages 20 to 40). The age ranges included in the personal training data are highlighted in red. We also provide an example image of the same person within 3 years of the target age as a reference.
}
    \label{fig:age_prog}
    \label{fig:age_reg}
\end{figure*} 

\clearpage


\bibliographystyle{ACM-Reference-Format}
\bibliography{references_new, jiaye, ref_update}

\clearpage
\setcounter{page}{1}
\maketitlesupplementary
\appendix
Along with this appendix, we provide additional visual materials (e.g., images and videos) on our project page~\footnote{~\href{https://mytimemachine.github.io/}{https://mytimemachine.github.io/}}.
We highly recommend viewing the accompanying videos for a more comprehensive look at the visual results.
\section{Overview of Appendices}
Our appendices contain the following additional details:
\begin{itemize} 
    \itemsep0em
    \item Sec.~\ref{sup:dataset} provides an overview of our data preprocessing pipeline, the curated dataset (summarized in Table~\ref{tab:dataset}), and additional details about the celebrities used in experiments Sec.~\ref{sec:compare}. 
    \item Sec.~\ref{sup:implementation_details} provides implementation details of our personalized adapter network, including hyperparameters and training configurations.
    \item Sec.~\ref{sup:compare_pretrained} presents benchmarking results against other pre-trained aging methods, with qualitative results shown in Fig.~\ref{fig:age_baseline}.
    \item Sec.~\ref{sup:compare_personalized} includes benchmarking results against alternative naive personalization techniques, with both quantitative and qualitative results displayed in Fig.~\ref{fig:ablation_personalization} and Fig.~\ref{fig:age_prog}.
    \item Sec.~\ref{sup:why_personalize_sam} explains our choice of using StyleGAN2's aging encoder for personalization over encoder-decoder GAN models or diffusion models.
    \item Sec.~\ref{sup:why_not_ref_for_swpaping} discusses the design rationale behind our video re-aging pipeline.
\end{itemize}

\section{Dataset Curation}
\label{sup:dataset}
Existing in-the-wild aging datasets~\cite{lin_fp-age_2022, ricanek_morph_2006, zhang_age_2017} lack longitudinal data for individual subjects, as they do not offer multiple high-quality images of the same person over several decades. 
To address this limitation, we followed previous personalization works~\cite{nitzan_mystyle_2022, zeng_mystyle_2023} and collected a new celeb dataset as summarized in Table~\ref{tab:dataset}. 
To better illustrate the age distributions, we count the number of images within the age ranges 20–40, 40–60, and 60–80, and report these numbers in the table. These ranges differ from the training ranges of 20–40, 50–70, and 30–70.
For each celebrity, we first gathered facial images, then enhanced older images to improve visual quality, compensating for the limitations of earlier camera technology and image processing methods. Following~\cite{wang_towards_2021}, we restored grayscale or low-quality images to ensure a more consistent and enhanced visual representation over time. Faces were then cropped and aligned according to the FFHQ~\cite{karras_style-based_2019} standard. While downloading publicly available images, we extracted metadata, such as the time of capture, to calculate each subject's age. 

\begin{table}[htbp]
\caption{
A longitudinal facial aging dataset featuring images of 12 celebrities. The number of images for each celebrity is reported across different age ranges. \textbf{Unless stated otherwise, 50 images are selected for training.}
}
\vspace{-.5em}
\centering
\resizebox{\columnwidth}{!}{%
\begin{tabular}{lcccc}
\toprule
\makecell[c]{Celebrity}
  & Age range & 20$\sim$40 & 40$\sim$60 & 60$\sim$80 
  \\
\midrule
Al Pacino & 21$\sim$84 & 89 & 56 & 198 \\
Charles III & 01$\sim$76 & 219 & 409 & 530 \\
Elizabeth II & 03$\sim$96 & 65 & 116 & 539 \\
Robert De Niro & 27$\sim$81 & 121 & 340 & 286 \\
Jennifer Aniston & 02$\sim$55 & 375 & 322 & - \\
Oprah Winfrey & 24$\sim$70 & 163 & 529 & 315 \\
Morgan Freeman & 20$\sim$87 & 4 & 136 & 290 \\
Jackie Chan & 21$\sim$70 & 31 & 444 & 201 \\
Chow Yun-fat & 20$\sim$68 & 91 & 109 & 60 \\
Elaine Chao & 16$\sim$71 & 14 & 117 & 123 \\
Diego Maradona & 17$\sim$60 & 165 & 301 & - \\
Margaret Thatcher & 20$\sim$87 & 70 & 270 & 268 \\
\bottomrule
\end{tabular}%
}
\label{tab:dataset}
\vspace{-2em}
\end{table}

\noindent
For re-aging tasks in Sec.~\ref{sec:compare}, the available age distribution of the collected celebrities varies; for instance, some celebrities have fewer than 50 images in the 20 to 40 age range. 
Therefore, we conduct age regression tasks for the following 10 celebrities: \textit{Al Pacino, Charles III, Elizabeth II, Robert De Niro, Oprah Winfrey, Morgan Freeman, Jackie Chan, Chow Yun-fat, Elaine Chao, and Margaret Thatcher}, and age progression tasks for the following 8 celebrities: \textit{Al Pacino, Charles III, Elizabeth II, Jennifer Aniston, Oprah Winfrey, Chow Yun-fat, Diego Maradona, and Margaret Thatcher}. These results also correspond to the number of pairs used in user studies, as discussed in Sec.~\ref{sec:compare}.
For dataset size ablation studies in Sec.~\ref{sec:abla}, we use the same celebrities selected in the regression task.

\vspace{-0.5em}
\section{Implementation Details}
\label{sup:implementation_details}
\textbf{Personalized Age Adapter Network.}
Inspired by~\cite{patashnik_styleclip_2021, bau_semantic_2019, liu_blendgan_2021}, our adapter network is built on a multi-layer perceptron (MLP) architecture that takes as input the latent vector \(\mathcal{W}_\text{tgt}^+\) and the target age ($a_\text{tgt}$), and outputs the offset vector \(\Delta{\mathcal{W}_\text{tgt}^+}\). Specifically, the $18 \times 512$ dimensional latent code \(\mathcal{W}_\text{tgt}^+\) is first processed through a Global MLP, which produces a down-sampled global representation $\mathcal{W}_\text{global}$ of dimension $18 \times 32$, flattened to $1 \times 512$. 
Next, we design an Aging MLP that takes the scalar target age as input and generates a $1 \times 16$ dimensional age feature, $a_{\text{tgt-feat}}$. We then train 18 independent Style MLPs, each operating on one of the $k \in [1,18]$ styles in the W+ space, to produce an offset vector for each style, $\Delta{\mathcal{W}_\text{tgt}^+(k)}$. Each Style MLP receives the $1 \times 512$ dimensional age-transformed latent code from SAM, $\mathcal{W}_\text{tgt}^+(k)$, the $1 \times 512$ dimensional global representation $\mathcal{W}_\text{global}$, and the target aging feature $a_{\text{tgt-feat}}$, and then outputs the per-style offset code $\Delta{\mathcal{W}_\text{tgt}^+(k)}$.
Both the Global, Aging, and 18 Style MLPs are designed as 2-layer neural networks with ReLU activation. This architecture enables the network to subtly and effectively adjust the latent representation, preserving the individual’s identity while incorporating personalized aging characteristics.

For each celebrity, we train our adapter network on a GPU A6000 for 10,000 iterations, which takes approximately 4 hours. We inherit SAM's hyperparameters, including its original loss weights. Additionally, we set $\lambda_\text{pers-age} = 1$ for Eq.~\ref{eq:pers_aging_loss}, $\lambda_\text{reg-extra} = 1$ for Eq.~\ref{eq:reg_extrapolation}, and  
$\lambda_\text{reg} = 1$ for Eq.~\ref{eq:reg_w_norm}.


\section{Comparison with SOTA Methods w/o Personalization}
\label{sup:compare_pretrained}
As discussed in Sec.~\ref{sec:compare}, we benchmark our approach against all available \colorbox{yellow}{open-sourced} pre-trained baselines, including SAM, CUSP, AgeTransGAN, and FADING, as shown in Fig.~\ref{fig:age_baseline}. We exclude RAGAN~\cite{makhmudkhujaev_re-aging_2021} and PADA~\cite{li_pluralistic_2023} from our comparisons as they are not open-sourced. For baseline methods like CUSP and AgeTransGAN, which utilize pre-defined age groups based on FFHQ-Aging~\cite{or-el_lifespan_2020}, we interpolate between these age groups to demonstrate continuous aging, following the approach used by SAM~\cite{alaluf_only_2021}. 

For benchmarking, we primarily focus on identity-preserving performance rather than inversion performance when the target age matches the input age. Consequently, reconstruction metrics such as PSNR are not included in our evaluation.

\section{Comparison with Naive Personalization Techniques.}
\label{sup:compare_personalized}
\begin{figure}[h]
\begin{minipage}{\columnwidth}
    \centering
    \includegraphics[width=\columnwidth]{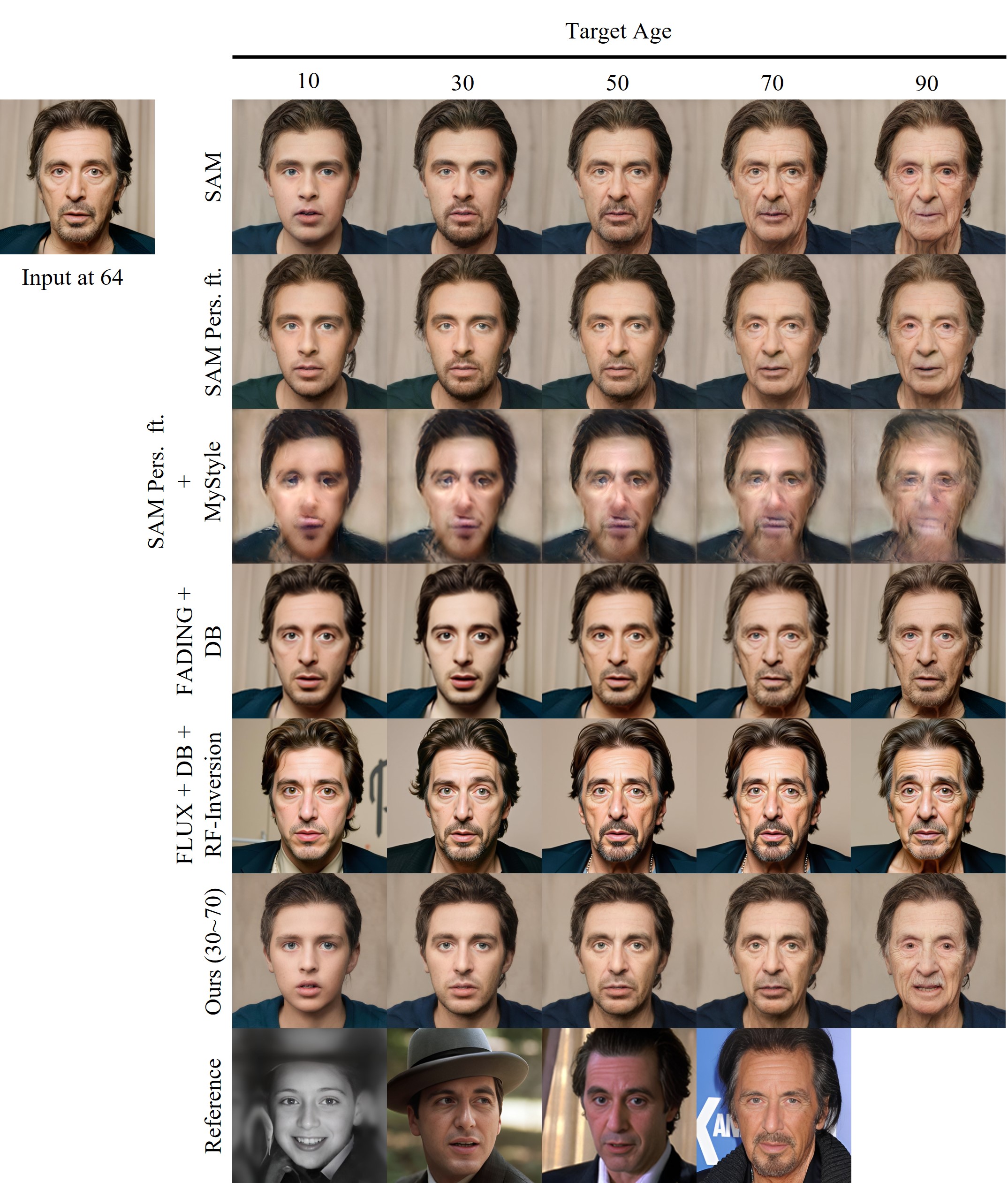}

    \vspace{2pt}

    \centering
    \resizebox{\columnwidth}{!}{%
        \begin{tabular}{ccccccc}
            \toprule
            Experiment & SAM & SAM Pers. ft. & \makecell[c]{SAM Pers. ft. + \\ MyStyle} & \makecell[c]{FADING + \\ DB} & 
            \makecell[c]{FLUX + DB + \\ RF-Inversion} & Ours (30$\sim$70) \\
            \midrule
            ID$_{sim} (\uparrow)$
            & 0.45
            & 0.49
            & 0.60
            & 0.64
            & 0.64
            & 0.66 \\
            \bottomrule
        \end{tabular}%
    }

    \vspace{-0.9em}
    
\caption{We compare MyTM (Ours) with naive personalization techniques: SAM Pers. ft., SAM Pers. ft. + MyStyle, FADING + Dreambooth (DB), and FLUX + Dreambooth (DB) + RF-Inversion, trained on ages 30$\sim$70 and tested within the same age range for \textit{Al Pacino}. While SAM Pers. ft. + MyStyle achieves a high $\text{ID}_{sim}$ score, it suffers from poor visual quality, resulting in adversarial examples for arcface.}

    \label{fig:ablation_personalization}
\end{minipage}
\vspace{-2em}
\end{figure}

We perform additional ablation studies using alternative personalization approaches on data for \textit{Al Pacino} aged $30\sim70$ years, with results shown in Fig.~\ref{fig:ablation_personalization}. SAM Pers. f.t. behaves similarly to the pre-trained SAM, as the latent codes are far from the latent center, limiting its editing capabilities. This aligns with the inversion-editability trade-off discussed in Sec.~\ref{sec:loss}. SAM Pers. ft. + MyStyle~\cite{nitzan_mystyle_2022} first personalizes the SAM encoder, then tunes the decoder following the PTI pipeline~\cite{roich_pivotal_2021}. However, this introduces significant artifacts due to changes in the latent distribution, which diverges from the pre-trained StyleGAN2 distribution. In SAM, global aging knowledge is learned with a fixed StyleGAN2 decoder, and modifying decoder weights distorts the latent space distribution, compromising the aging knowledge and introducing decoding artifacts. 

For naive personalization using diffusion models, 
FADING + Dreambooth (DB)~\cite{ruiz_dreambooth_2023} overfits the aging results to the input image, especially when the target age lies outside the training age range. Additionally, this approach neglects age-related facial shape transformations, such as a toddler’s rounder face or proportional changes in facial features over time, which are caused by NTI + p2p as discussed in Sec.~\ref{sec:related_work}. 
We further adopt FADING’s paradigm using a more contemporary pipeline: FLUX~\footnote{https://huggingface.co/black-forest-labs/FLUX.1-dev} combined with RF-Inversion~\cite{rout2025semantic} and DreamBooth (DB), replacing the original Stable Diffusion 1.5 + NTI setup in FADING + DB. However, this updated pipeline still fails to extrapolate effectively, exhibiting similar overfitting failures as FADING + DB.

\begin{figure*}[!t]
    \centering
    \vspace{-0.3em}
    \includegraphics[width=\linewidth]{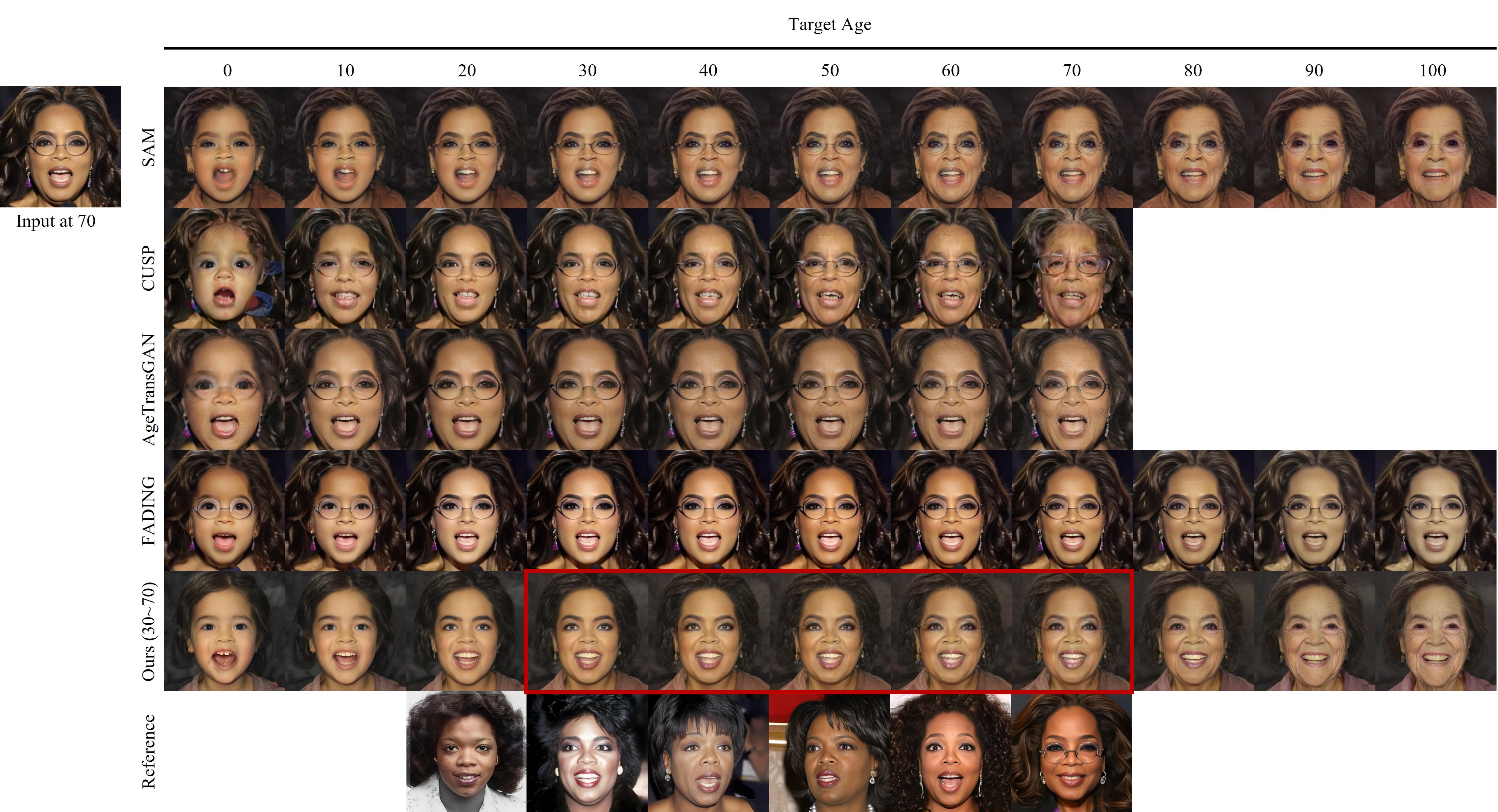}
\caption{Limitations of MyTM. Our method may struggle with accessories (e.g., glasses), as these elements are not consistently handled by the e4e encoder~\cite{tov_designing_2021}.}
    \label{fig:age_regression_oprah}
\end{figure*} 

\section{Why Personalizing the Encoder SAM?}
\label{sup:why_personalize_sam}
Finetuning encoder-decoder GAN with limited personal data often leads to overfitting, mode collapse~\cite{pmlr-v202-aghabozorgi23a}, and data drift~\cite{lee_countering_2019, lu_countering_2020}, preventing the model
from generalizing to unseen test images of an individual~\cite{qi_my3dgen_2023}.  
Therefore, encoder-decoder GAN structures, like AgeTransGAN, necessitate a substantial amount of paired data to achieve effective personalization in aging transformations. For instance, personalizing the appearance of a celebrity such as \textit{Al Pacino} would necessitate images of him at both ages 20 and 70, with consistent pose, lighting, and expression. However, acquiring such data in real-world conditions is extremely challenging, as it demands rare and specific longitudinal images that capture individuals across a wide age span under controlled settings. This limitation makes encoder-decoder GANs less practical for applications where personalized aging transformations are desired.

\begin{figure*}[th]
    \centering
    \includegraphics[width=\linewidth]{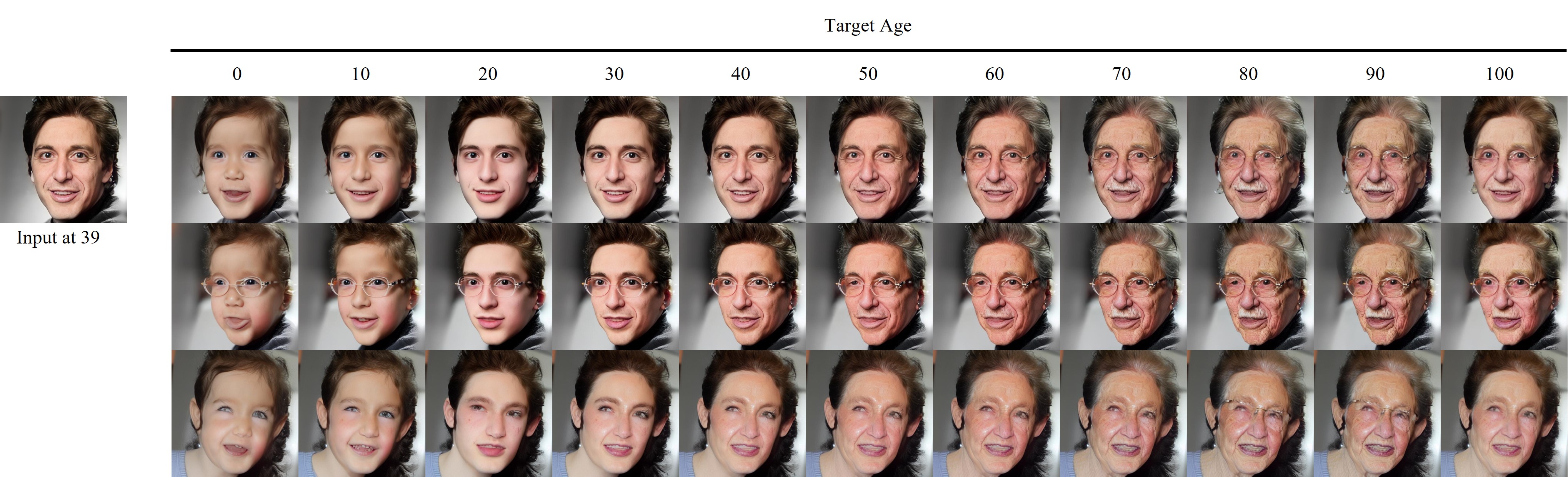}
    \caption{Visual results of FADING using identical input and inference code. The instability in age transformation arises from the optimization of NTI~\cite{mokady_null-text_2022}, leading to inconsistencies.}
    \label{fig:fading_failure}
\end{figure*}

\noindent
For diffusion models, there are several limitations in re-aging tasks: (1) They lack the W latent space, which enables fine-grained continuous aging control and editing~\cite{dravid_interpreting_2024}. (2) Models like FADING, which use NTI + p2p for age editing, often struggle with the trade-off between inversion accuracy and editability~\cite{rout_semantic_2024}. Additionally, FADING frequently produces unstable results, as shown in Fig.~\ref{fig:fading_failure}, which we attribute to the unstable NTI optimization~\cite{zheng_inversemeetinsert_2024, rout_semantic_2024}. New stable optimization-free methods could be explored for diffusion models in the future. (3) VQ auto-encoders, commonly used in diffusion to encode images, can introduce artifacts, particularly in the human face domain~\cite{mokady_null-text_2022}. These issues highlight the need for an alternative approach, such as utilizing StyleGAN2's well-trained latent space and optimization-free e4e encoder~\cite{tov_designing_2021}, to achieve high-quality, artifact-free re-aging transformations.

\section{Why Not Use a Reference Image for Face-Swapping?}
\label{sup:why_not_ref_for_swpaping}
Firstly, obtaining images of a person at any arbitrary age is often challenging, particularly high-quality images comparable to our synthesized faces at 1024x1024 resolution. Even if reference images at the target age are available, face-swapping techniques~\cite{chen_simswap_2021, xu_designing_2022, gao_information_2021} generally yields optimal results when the source and target faces share similar styles, such as pose, expression, and lighting. Significant style differences between the source and target faces can cause artifacts like flickering, particularly in real-world video scenarios~\cite{ren_reinforced_2023, choi_personalized_2023}.










\end{document}